\documentclass{article} 
\usepackage{iclr2024_conference,times}

\usepackage{amsmath,amsfonts,bm}

\def\eqref#1{equation~\ref{#1}}

\def\1{\bm{1}}

\DeclareMathAlphabet{\mathsfit}{\encodingdefault}{\sfdefault}{m}{sl}
\SetMathAlphabet{\mathsfit}{bold}{\encodingdefault}{\sfdefault}{bx}{n}

\usepackage{hyperref}
\usepackage{url}

\usepackage{subcaption}

\usepackage{wrapfig}
\usepackage{lipsum}
\usepackage[utf8]{inputenc} 
\usepackage[T1]{fontenc}    
\usepackage{hyperref}       
\usepackage{url}            
\usepackage{booktabs}       
\usepackage{nicefrac}       
\usepackage{microtype}      
\usepackage{xcolor}         
\usepackage{multirow}
\usepackage{tabularx}
\usepackage{threeparttable}
\usepackage{enumitem}
\usepackage{algorithm}
\usepackage{setspace}
\usepackage{algorithmicx}
\usepackage{latexsym}
\usepackage{graphicx}
\usepackage{amsmath}
\usepackage{algorithm}
\usepackage{algorithmicx}
\usepackage{algpseudocode}
\usepackage{multirow}
\usepackage{tabularx}
\usepackage{amsfonts}  
\usepackage{threeparttable}
\usepackage{enumitem}
\usepackage{amsmath}
\usepackage{graphicx}
\usepackage{multirow}
\usepackage{threeparttable}
\usepackage{tabularx}
\usepackage{algorithm}
\usepackage{setspace}
\usepackage{hyperref} 
\usepackage{xcolor}
\usepackage{soul}
\usepackage{cleveref}
\usepackage{caption}
\usepackage{dashbox}
\newtheorem{assumption}{Assumption}

\definecolor{green1}{RGB}{5,111,191}
\DeclareRobustCommand{\hlgreen1}[1]{{\sethlcolor{green1}\hl{#1}}}
\title{Towards Faithful Explanations: Boosting Rationalization with Shortcuts Discovery}
\author{Linan Yue$^{1}$, Qi Liu$^{1,2}$\thanks{Corresponding author.}\, , Yichao Du$^{1}$, Li Wang$^{3}$, Weibo Gao$^{1}$, Yanqing An$^{1}$\\
1: State Key Laboratory of Cognitive Intelligence, \\University of Science and Technology of China\\
2: Institute of Artificial Intelligence, Hefei Comprehensive National Science Center Hefei, China\\
3: ByteDance\\
\texttt{\{lnyue,duyichao,wl063,weibogao,anyq\}@mail.ustc.edu.cn}; \\
\texttt{qiliuql@ustc.edu.cn} 
}

\iclrfinalcopy 
\begin{document}

\maketitle
\begin{abstract}
   The remarkable success in neural networks provokes the selective rationalization. It explains the prediction results by identifying a small subset of the inputs sufficient to support them.
   Since existing methods still suffer from adopting the shortcuts in data to compose rationales and limited large-scale annotated rationales by human, in this paper, we propose a Shortcuts-fused Selective Rationalization (SSR) method, which boosts the rationalization by discovering and exploiting potential shortcuts.
   Specifically, SSR first designs a shortcuts discovery approach to detect several potential shortcuts.
   Then, by introducing the identified shortcuts, we propose two strategies to mitigate the problem of utilizing shortcuts to compose rationales.
   Finally, we develop two data augmentations methods to close the gap in the number of annotated rationales.
   Extensive experimental results on real-world datasets clearly validate the effectiveness of our proposed method. Code is released at \url{https://github.com/yuelinan/codes-of-SSR}.
   \end{abstract}
   
   \section{Introduction}
\label{intro}
Although deep neural networks (DNNs) in natural language understanding tasks have achieved  compelling success, their predicted results are still unexplainable and unreliable, prompting significant research into how to provide explanations for DNNs.
Among them, the selective rationalization \citep{lei-etal-2016-rationalizing,bastings-etal-2019-interpretable,paranjape-etal-2020-information,li2022unifying} has received increasing attention, answering ``What part of the input drives DNNs to yield prediction results?''.
Commonly, the rationalization framework consists of a \textit{selector} and a \textit{predictor}. 
The goal of rationalization is to yield task results with the \textit{predictor}, while employing the \textit{selector} to identify a short and coherent part of original inputs (i.e., rationale), which can be sufficient to explain and support the prediction~results. 

\begin{figure}[!ht]
  \centering
  \includegraphics[width=7.8cm]{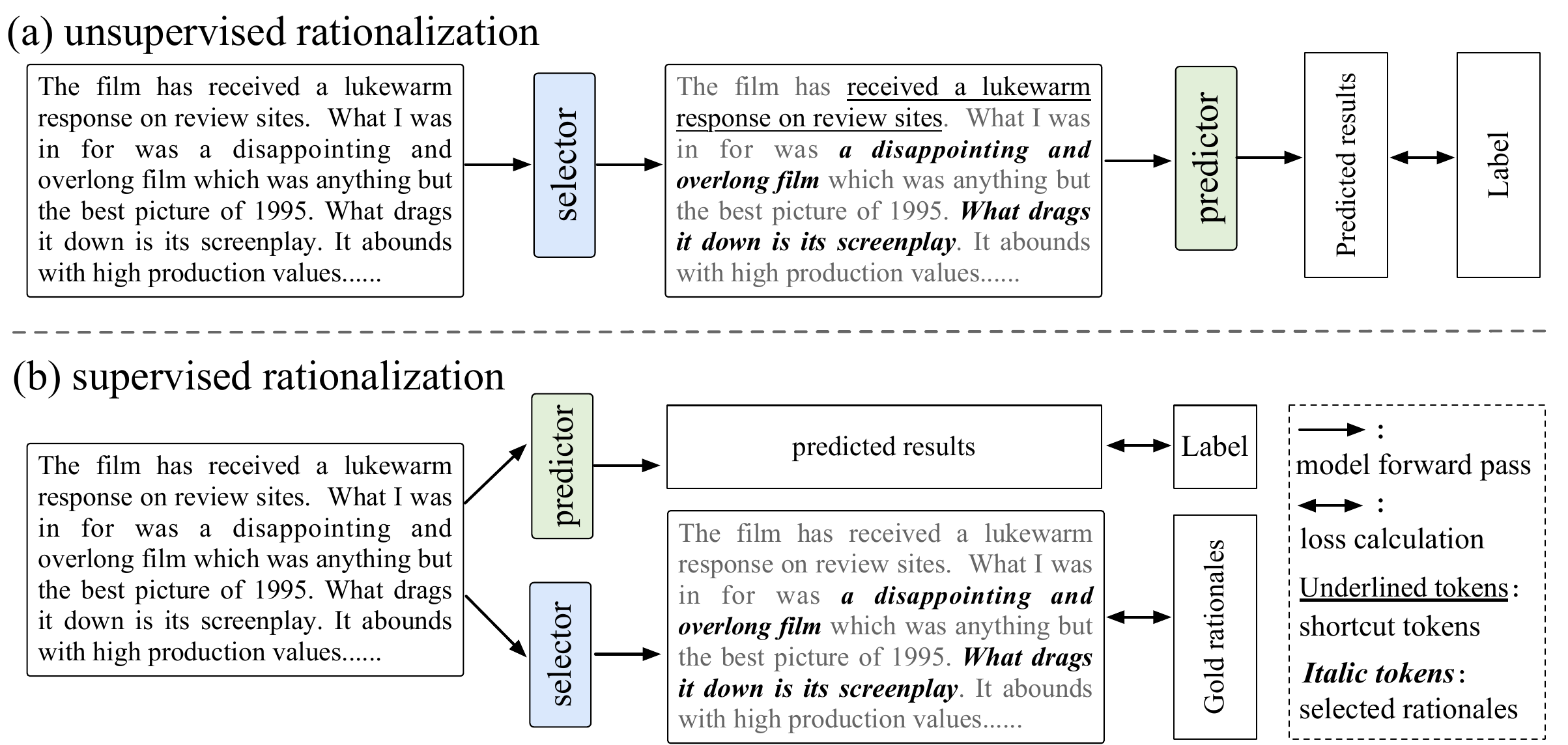}
   \vspace{-0.3cm}
  \caption{
      Schematic of rationalization methods presented in this paper.
      (a) is the process of unsupervised rationalization with the \textit{selector}-\textit{predictor} pattern.
      (b) illustrates the supervised rationalization with a multi-task framework. 
      Semi-rationalization can be considered the combination of (a) and (b).
  } 
  \label{example}
   \vspace{-0.7cm}
  \end{figure}

Existing selective rationalization methods can be grouped into three types.
The first type trains the \textit{selector} and  \textit{predictor} in tandem \citep{lei-etal-2016-rationalizing,bastings-etal-2019-interpretable,paranjape-etal-2020-information}.
Specifically, as shown in Figure~\hyperref[example]{1(a)}, it adopts the \textit{selector} to extract a text span from the input (i.e., rationale), and then yields the prediction results solely based on the selective text by the \textit{predictor}.
It is worth noting that the gold rationale is unavailable during the whole training process.
Therefore, we refer to this type of method as ``\textit{unsupervised rationalization}''.
Although this approach achieves promising results, recent studies \citep{chang2020invariant,wu2022dir} have proved that the success of this method is exploiting the shortcuts in data to make predictions.
Typically, shortcuts have potentially strong correlations (aka., spurious correlations) with task labels, but would not be identified as rationales to the prediction task by human.
For instance, in Figure~\hyperref[example]{1(a)}, there is a movie review example whose label is ``\textit{negative}''.
Among them, the underlined tokens extracted by the unsupervised rationalization method are the shortcut tokens, where a poor quality movie is always associated with ``received a lukewarm response''. An unsupervised rationalization is easy to predict the movie as ``\textit{negative}'' based on  these shortcut tokens.
However, for a human being, judging a movie is not influenced by other movie reviews (i.e., even if a movie has a low rating on movie review sites, someone will still enjoy it). Therefore, although the model predicts the right outcome, it still fails to reveal true rationales for predicting labels but depending on the shortcuts. In general, while shortcuts are potentially effective in predicting task results, they are still damaging to compose rationales.
Based on this conclusion, we can get an important assumption:
\textcolor{black}{\assumption{
  \label{assumption1}
{A well-trained unsupervised rationalization model inevitably composes rationales with both the gold rationale and shortcuts tokens.}}}


Then, as the second type is to include the gold rationales annotated by human during training \citep{DeYoungJRLXSW20,li2022unifying,chan2022unirex}, we denote it as  ``\textit{supervised rationalization}''. It models the rationalization with a multi-task learning, optimizing the joint likelihood of class labels and extractive rationales (Figure \hyperref[example]{1(b)}).
Among them, the rationale prediction task can be considered as a token classification.
Since this method exploits real rationales, the problem of adopting the shortcuts to predict task results can be mitigated.
However, such extensive annotated rationales are infeasible to obtain for most tasks, rendering this method~unavailable.

To combine the superiority of the above two types of methods, \cite{pruthi-etal-2020-weakly,bhat2021self} propose a  ``\textit{semi-supervised rationalization}'' method, consisting of a \textit{two-phases} training.~They~first train the rationalization task with few labeled rationales in a multi-task learning framework (\textit{the supervised phase}) like the second type method, where we denote the used training dataset as~$\mathcal{D}_{sup}$.
Then, they train the remaining data $\mathcal{D}_{un}$ following the first type method (\textit{the unsupervised phase}).
Since this method still suffers from limited gold rationales and employing shortcuts to generate rationales, we argue this ``\textit{semi-supervised pattern}'' can be further explored to improve the rationalization.

Along this research line, in this paper, we propose a boosted method \textbf{S}hortcuts-fused \textbf{S}elective \textbf{R}ationalization (SSR) which enhances the semi-supervised rationalization by exploring shortcuts.
Different from the previous methods that are degraded by shortcuts, SSR 
explicitly exploits shortcuts in the data to yield more accurate task results and extract more plausible rationales.
Specifically, in the semi-supervised setting, we first train SSR with $\mathcal{D}_{sup}$ in \textit{\underline{the supervised phase}}.
Among them, since there exist no labeled shortcuts, we design a shortcuts discovery approach to identify several potential shortcut tokens in $\mathcal{D}_{sup}$.
In detail, we employ a trained \textit{unsupervised rationalization} model to infer potential rationales in $\mathcal{D}_{sup}$.
As discussed in \textbf{Assumption \ref{assumption1}}, the rationales extracted by \textit{unsupervised rationalization} inevitably contain several shortcut tokens.
Then, by introducing the gold rationales, we can explicitly obtain the shortcut tokens.
Next, we design two strategies to learn the extracted shortcut information and further transfer it into \textit{\underline{the unsupervised phase}}, which can mitigate the problem of adopting shortcuts to yield rationales.
Besides, due to the limited rationales labels, we develop two data augmentations methods, including a random data augmentation and semantic augmentation, by replacing identified shortcut tokens in $\mathcal{D}_{sup}$.
To validate the effectiveness of our proposed SSR, we conduct extensive experiments on five datasets from the ERASER benchmark~\citep{DeYoungJRLXSW20}.
The experimental results empirically show that SSR consistently outperforms the competitive unsupervised and semi-supervised baselines on both the task prediction and rationale generation  by a significant margin, and achieves comparable results to supervised baselines.


\section{Problem Formulation}

Considering a text classification task, given the text input $x=\left\{x_{1}, x_{2}, \ldots, x_{n}\right\}$  and the ground truth~$y$, where $x_{i}$ represents the $i$-th token, the goal is employing a \textit{predictor} to yield the prediction results~${y}$ while learning a mask variable $m=\left\{m_{1}, m_{2}, \ldots, m_{n}\right\}$ by a \textit{selector}.
Among them, $m_{j} \in\{0,1\} $ indicates whether the $i$-th token is a part of the rationale. Then, the selected rationale is defined as 
$z = m \odot x= \left\{m_{1}\cdot x_{1}, m_{2} \cdot x_{2}, \ldots, m_{n} \cdot x_{n} \right\}$.
Take the case in Figure \ref{example} for example, the selective rationalization aims to yield an accurate prediction result (i.e., \textit{negative}) and extract the rationale (the italic tokens) as the supporting evidence to explain this result.

\section{Preliminary of Selective Rationalization}
\textbf{Unsupervised rationalization.}
\label{2.2}
In the unsupervised rationalization, since the gold rationales are unavailable, to achieve extracting rationales, this type of method trains the \textit{selector} and \textit{predictor} in tandem. Specifically, the \textit{selector} first maps each token $x_{i}$ to its probability, $p_{\theta}(\widetilde{m}_{i}|x_{i})$ of being selected as part of rationale, where $p_{\theta}(\widetilde{m}_{i}|x_{i}) = \textrm{softmax}(W_{s_{un}}f_{s_{un}}(x_{i}))$.
Among them, $f_{s_{un}}(\cdot)$ represents an encoder (e.g. BERT~\citep{devlin-etal-2019-bert}), encoding $x_{i}$ into a $d$-dimensional vector, and ${W}_{s_{un}} \in \mathbb{R}^{2 \times d}$.
Then, to sample $m_{i} \in\{0,1\}$ from the $p_{\theta}(\widetilde{m}_{i}|x_{i})$ distribution and ensure this operation differentiable, \cite{lei-etal-2016-rationalizing} introduce a Bernoulli distribution with REINFORCE \citep{williams1992simple}. Since this method may be quite unstable \citep{paranjape-etal-2020-information}, in this paper, we implement this sampling operation with a Gumbel-Softmax reparameterization \citep{jang2017categorical}: 
\begin{equation}
    m_{i}=\frac{\exp \left(\left(\log \left(p_{\theta}(\widetilde{m}_{i}|x_{i})\right)+g_{i}\right) / \tau\right)}{\sum_{j} \exp \left(\left(\log \left(p_{\theta}(\widetilde{m}_{j}|x_{j})\right)+g_{j}\right) / \tau\right)} ,
\end{equation}
where $g_{i}=-\log \left(-\log \left(u_{i}\right)\right)$ and $\tau$ is a temperature hyperparameter. $u_{i}$ is sampled from a uniform distribution $U(0,1)$.
Details of Gumbel-Softmax are shown in Appendix \ref{gumbel}.
Naturally, the rationale $z$ extracted by the \textit{selector} is calculated as $z = m \odot x$.

Next, the \textit{predictor} $q_{\psi}(y|z)$ yields the prediction results solely based on the rationale $z$, where $q_{\psi}(y|z) = \textrm{softmax}(W_{p_{un}}f_{p_{un}}(z))$. Wherein $f_{p_{un}}(\cdot)$ re-encodes $z$ into $d$-dimensional continuous hidden states, $W_{p_{un}} \in \mathbb{R}^{N \times d}$ is the learned parameters, and $N$ is the number of labels (e.g., $N=2$ in the binary classification). Finally, the prediction loss can be formulated as :
\begin{equation}
    \mathcal{L}_{un\_task}  = \mathbb{E}_{\substack{{x}, y \sim \mathcal{D}_{un} \\ {m} \sim p_{\theta}({\widetilde{m}} | x)}}\left[-\log q_{\psi}(y | m \odot x)\right],
    \label{eqloss}
\end{equation}
where $\mathcal{D}_{un}$ is a training set (gold rationales are unavailable).
Besides, to impose the selected rationales are short and coherent, we incorporate sparsity and continuity constraints into the rationalization:
\begin{equation}
    \mathcal{L}_{re} =  \underbrace{ \lambda_{1} \left| \frac{1}{n}\sum_{i=1}^{n}m_{i} - \alpha \right|}_{sparsity}+ \underbrace{\lambda_{2}\sum_{i=2}^{n}\left|{m}_{i}-{m}_{i-1}\right|}_{continuity},
    \label{sparsity}
\end{equation}
where $\alpha \in [0,1] $ is the predefined sparsity level (the higher the $\alpha$, the lower the sparsity).
Finally, the  objective of the unsupervised rationalization is defined as $\mathcal{L}_{un} = \mathcal{L}_{un\_task} + \mathbb{E}_{\substack{ {m}\sim p_{\theta}({\widetilde{m}} | x)}}  \left[\mathcal{L}_{re}\right]$.

Although this type of method can extract rationales without the labeled rationales supervision and achieve promising results \citep{bastings-etal-2019-interpretable,sha2021learning,yu2021understanding}, \cite{chang2020invariant,wu2022dir} have proved that it is prone to exploiting the shortcuts in data (e.g., the statistics shortcuts) to yield prediction results and rationales.
In other words, such shortcut-involved rationales fail to reveal the underlying relationship between inputs and rationales.

\textbf{Supervised rationalization.}
\label{2.3}
Since both the ground truth and the human annotated rationales are available in supervised rationalization, several researches \citep{li2022unifying,chan2022unirex} introduce a joint task classification and rationalization method. 
Specifically, in the supervised rationalization, we denote $\mathcal{D}_{sup}$ as a joint task and rationale labeled training set, containing additional annotated rationales $\hat{z}$.
Then, similar to the unsupervised rationalization, we generate the probability of selecting $x_{i}$ as the part of rationales by calculating $p_{\theta}(\widetilde{m}_{i}|x_{i})=\textrm{softmax}(W_{s_{sup}}f_{s_{sup}}(x_{i}))$. Next, given gold rationales $\hat{z}$, we can consider the rationalization task as a binary token classification, and calculate the corresponding loss with token-level binary cross-entropy (BCE) criterion :
\begin{equation}
    \mathcal{L}_{select}  = \sum_{i=1}^{n}-\hat{m}_{i} \log p_{\theta}(\widetilde{m}_{i}|x_{i}),
\end{equation}
where $\hat{m}=\left\{\hat{m}_{1}, \hat{m}_{2}, \ldots, \hat{m}_{n}\right\}$ is the gold mask corresponding to $\hat{z}$.
Next, for the task classification, different from the unsupervised rationalization employing extracted rationales as the input, the \textit{predictor} in supervised rationalization yields results with $q_{\psi}(y|x)$, and the prediction loss is calculated as $\mathcal{L}_{sup\_task}  = \mathbb{E}_{\substack{{x}, y \sim \mathcal{D}_{sup}}}\left[-\log q_{\psi}(y | x)\right]$, where $q_{\psi}(y | x)=\textrm{softmax}(W_{p_{sup}}f_{p_{sup}}(x))$.
Finally, the objective of supervised rationalization is $\mathcal{L}_{sup} = \mathcal{L}_{sup\_task} + \mathbb{E}_{\substack{ {x}, z  \sim \mathcal{D}_{sup}}}  \left[\mathcal{L}_{select}\right]$.

Since the real rationale label is explicitly introduced into the supervised rationalization, 
the ``shortcuts'' problem posed by unsupervised methods can be eased.
However, such extensive rationales annotated by human are the main bottleneck enabling the widespread application of these models.

\textbf{Semi-supervised rationalization.}
\label{2.4}
For the semi-supervised rationalization, researchers \citep{paranjape-etal-2020-information,pruthi-etal-2020-weakly,bhat2021self} consider a low-resource setup where they have annotated rationales for part of the training data $\mathcal{D}_{semi}$.
In other words, $\mathcal{D}_{semi}$ consists of $\mathcal{D}_{un}$ and $\mathcal{D}_{sup}$, where $|\mathcal{D}_{un}| \gg |\mathcal{D}_{sup}|$. 
Then, we use the following semi-supervised objective: $\mathcal{L}_{semi} =  \mathcal{L}_{sup} + \mathcal{L}_{un}$, where we first train the model on $\mathcal{D}_{sup}$ and then on $\mathcal{D}_{un}$.

Despite this method appearing to combine the advantages of the previous two methods, it is still prone to adopting shortcuts for prediction ($\mathcal{L}_{un}$ in $\mathcal{L}_{semi}$).
Besides, due to the gap in data size between the two datasets (i.e., $\mathcal{D}_{un}$ and $\mathcal{D}_{sup}$), the prediction performance of this semi-supervised approach is considerably degraded compared to the supervised rationalization \citep{bhat2021self}.
\vspace{-0.4cm}
\section{Shortcuts-fused~Selective~Rationalization}
\vspace{-0.3cm}
In this section, we follow the above semi-supervised rationalization framework and further propose a {S}hortcuts-fused {S}elective {R}ationalization (SSR) method by discovering shortcuts in data. 
We first identify shortcut tokens from the input by exploring gold rationales in $\mathcal{D}_{sup}$ (section~\ref{ssr3.1}).
Then, to mitigate the problem which exploits shortcuts for prediction, we introduce two strategies by leveraging identified shortcuts (section \ref{ssr3.2}).
Finally, to bridge the data size gap between $\mathcal{D}_{un}$ and $\mathcal{D}_{sup}$, we develop two data augmentation methods also depending on identified~shortcuts  (section~\ref{ssr3.3}).

\subsection{Shortcuts Discovery}

\label{ssr3.1}
\begin{wrapfigure}[9]{r}{7.2cm}
  \centering
    \vspace{-1.2cm}
  \setlength{\abovecaptionskip}{0.cm}
  \includegraphics[width = 6.8cm]{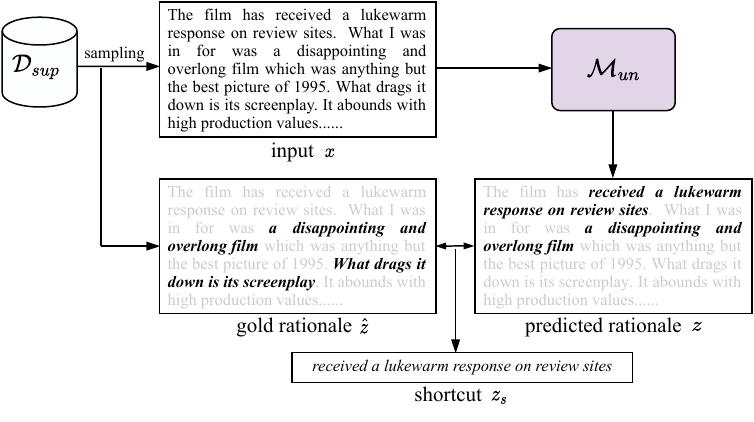}
  \setlength{\abovecaptionskip}{0.cm}
  \caption{
    Process of the shortcut generator.
  }
  \label{shortcut}
\end{wrapfigure}
\label{3.1}

As described above, identifying and detecting shortcuts in the data is fundamental to our approach. However, since there exist no labeled shortcuts, posing a challenge to discover shortcuts.
Based on \textbf{Assumption \ref{assumption1}}, we can identify the potential shortcut token from   rationales extracted by a well-trained unsupervised rationalization model
via introducing labeled rationale tokens.

\textbf{Definition 1} 
\textit{ (Potential Shortcut Token) We first assume the unsupervised rationalization model $\mathcal{M}_{un}$ is already trained. Then, given the annotated rationales $\hat{z}$ and rationales $z$ extracted by  $\mathcal{M}_{un}$, we define  
$\mathbb{PST}(x_{i})$ as  whether a token $x_{i}$ is considered to be  a potential shortcut token or not:
\begin{equation}
  \mathbb{PST}(x_{i}) = \mathbb{I}(x_{i} \in {z} \wedge x_{i} \notin \hat{z}),
\end{equation}
where $\wedge$ is the logical operation AND. $\mathbb{PST}(x_{i})$$=$$1$ denotes $x_{i}$ is defined as a potential shortcut~token.}

Motivated by the above assumption and definition, our shortcuts discovery method is two-fold:

\textbf{(\textit{i})} We train an unsupervised rationalization method~$\mathcal{M}_{un}$ with $\mathcal{D}_{un}$, which sufficiently exploits shortcuts to make predictions.

\textbf{(\textit{ii})} We design a shortcut generator (Figure \ref{shortcut}) to combine labeled rationales in $\mathcal{D}_{sup}$ and $\mathcal{M}_{un}$ to identify shortcuts. Specifically, the shortcut generator first employs $\mathcal{M}_{un}$ to infer the potential rationales $z$ in $\mathcal{D}_{sup}$. Next, we introduce the gold rationales $\hat{z}$ and compare them with the predicted rationales $z$. If a token $x_{i} \in {z}$ and $x_{i} \notin \hat{z}$ (i.e., this token is incorrectly predicted as rationale tokens), we define it as a potential shortcut token.

\begin{figure}[!htp]
  \centering
  \includegraphics[width=13cm]{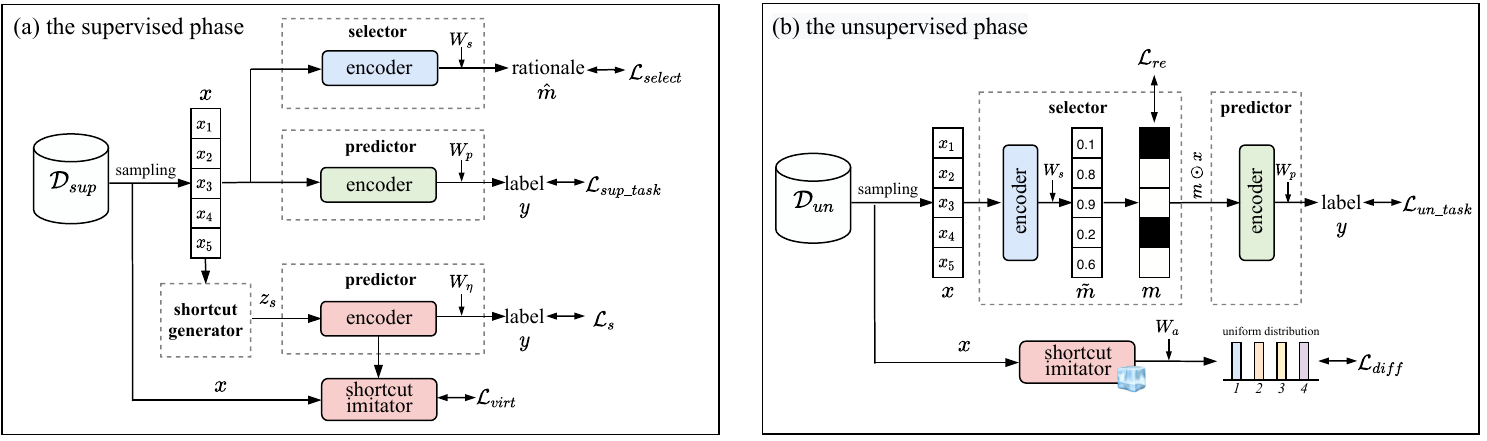}
  \caption{
      Architecture of $\textmd{SSR}_{virt}$ consisting of the supervised and unsupervised phases.
       Among them, \protect\includegraphics[width=8pt]{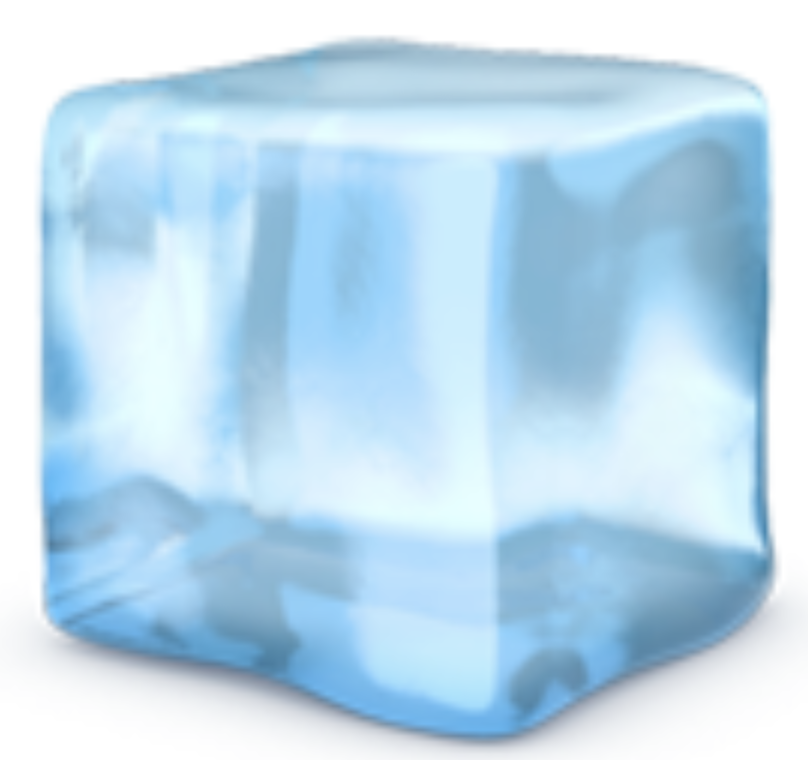} represents the frozen shortcut imitator, and white boxes in $m$ indicate the rationale tokens and the black are non-rationale ones.
  } 
  \label{model}
  \vspace{-0.5cm}
\end{figure}
In the practical implementation, considering the coherence of rationales, we identify a subsequence with three or more consecutive potential shortcut tokens as the shortcut $z_{s}$.
For example, as shown in Figure \ref{shortcut}, $\mathcal{M}_{un}$ has predicted the movie review as ``\textit{negative}'' correctly and composed the predicted rationales.
By comparing with  gold rationales, we identify ``\textit{received a  lukewarm response on review sites}'' as shortcuts. It's worth noting that the shortcut discovery is only used in the training phase.

\subsection{Two Strategies by Exploring Shortcuts}
\label{ssr3.2}
As mentioned previously, we follow the semi-rationalization framework and train SSR on $\mathcal{D}_{un}$ and $\mathcal{D}_{sup}$ in tandem.
Since we have identified the shortcuts in $\mathcal{D}_{sup}$, we design two strategies by exploring identified shortcuts to mitigate the impact of adopting shortcuts for prediction in $\mathcal{D}_{un}$.
\subsubsection{Shared Parameters.}
Before we introduce the two strategies, we present the shared parameters in both unsupervised and supervised rationalization.
Here, we let ``$A \leftrightarrow B$''  represent the parameter $A$ and $B$ share parameters.

\noindent$\bullet$ To allow the model to capture richer interactions between task prediction and rationales selection, we adopt $f_{s_{un}}(\cdot) \leftrightarrow f_{p_{un}}(\cdot)$, and  $f_{s_{sup}}(\cdot) \leftrightarrow f_{p_{sup}}(\cdot)$ \citep{bhat2021self,liu2022fr}.

\noindent$\bullet$ To let the unsupervised and supervised rationalization learn better from each other, we let $W_{s_{un}}f_{s_{un}}(\cdot) \leftrightarrow W_{s_{sup}}f_{s_{sup}}(\cdot)$  in the \textit{selector}. $W_{p_{un}}f_{p_{un}}(\cdot)\leftrightarrow W_{p_{sup}}f_{p_{sup}}(\cdot)$ in the \textit{predictor}.

In summary, all encoders in both \textit{selector} \underline{and} \textit{predictor} share parameters:
$f_{s_{un}}(\cdot) \leftrightarrow f_{p_{un}}(\cdot)  \leftrightarrow  f_{s_{sup}}(\cdot) \leftrightarrow f_{p_{sup}}(\cdot) $.
Linear parameters in \textit{selector} are shared: $W_{s_{un}} \leftrightarrow W_{s_{sup}}$.
Linear parameters in \textit{predictor} are shared: $W_{p_{un}} \leftrightarrow W_{p_{sup}}$.
For clarity, in both unsupervised and supervised rationalization, we represent the 
$f_{s}(\cdot)$ and $f_{p}(\cdot)$ as the encoder in \textit{selector} and  \textit{predictor}, $W_{s}$ and  $W_{p}$ are the corresponding linear parameters.
Table \ref{share-p} in Appendix \ref{app_share} lists all shared parameters.

\subsubsection{Injecting Shortcuts into Prediction.}
\label{3.2.2}
In the semi rationalization, the unsupervised rationalization may still identify the shortcuts as rationales due to the unavoidable limitations (section \ref{2.2}).
To this end, we propose a strategy, injecting shortcuts into the task prediction.
Specifically:

\textit{\textbf{In the supervised phase}}, besides the original loss $\mathcal{L}_{sup}$, we add a ``\textit{uniform}'' constraint to ensure the \textit{predictor} $q_{\psi}$ identifies the shortcuts features as meaningless features:
\begin{equation}
    \mathcal{L}_{unif}=\mathbb{E}_{\substack{ z_{s}  \sim \mathcal{D}_{sup}}}\left[\operatorname{KL}\left( \mathcal{U}(0, |N|) \|  q_{\psi}(y|z_{s})\right)\right],
    \label{kl}
\end{equation}
where KL denotes the Kullback–Leibler divergence, $|N|$ is the total number of classes, and $\mathcal{U}(0, |N|)$ denotes the uniform class distribution.
When the \textit{predictor} $q_{\psi}(y|x)$ adopts the input to yield task results,
Eq~(\ref{kl}) encourages the \textit{predictor} to identify shortcut tokens $z_{s}$ as meaningless tokens, and disentangles shortcuts features from the input ones (i.e., making shortcuts and rationales de-correlated).

\textit{\textbf{In the unsupervised phase}}, the learned features information can be transferred into the unsupervised rationalization method through the shared \textit{predictor} ($W_{p_{un}}f_{p_{un}}(\cdot)\leftrightarrow W_{p_{sup}}f_{p_{sup}}(\cdot)$).

Finally, we denote SSR with this  strategy as  $\textmd{SSR}_{unif}$, and
the objective of $\textmd{SSR}_{unif}$ can be defined as the sum of the losses:
$\mathcal{L}_{ssr_{unif}} = \mathcal{L}_{un} + \mathcal{L}_{sup} + \lambda_{unif} \mathcal{L}_{unif}$.
Detailed algorithms of $\textmd{SSR}_{unif}$ are shown in Appendix \ref{A.1.algo}.

\begin{table*}[htbp]
  \caption{Task F1 and Token F1 of selected rationales for the five datasets. Among them, the underlined scores are the state-of-the-art performances of the supervised rationalization.}
  \renewcommand\arraystretch{.4}
  \setlength{\tabcolsep}{2.73mm}{
      \scalebox{.58}{
          \begin{tabular}{c|cc|cc|cc|cc|cc}
              \hline    
              \toprule
                  \multirow{2}{*}{Methods} &\multicolumn{2}{c|}{Movies} & \multicolumn{2}{c|}{MultiRC} & \multicolumn{2}{c|}{BoolQ} &\multicolumn{2}{c}{Evidence Inference} & \multicolumn{2}{c}{FEVER}\\\cmidrule(r){2-11}
                  &Task&Token-F1&Task&Token-F1&Task&Token-F1 &Task&Token-F1&Task&Token-F1 \\ 
              \midrule
              Vanilla Un-RAT  &87.0 $\pm$  0.1  & 28.1 $\pm$  0.2 & 57.7 $\pm$  0.4 & 23.9 $\pm$  0.5 & 62.0 $\pm$  0.2 & 19.7 $\pm$  0.4 & 46.2 $\pm$  0.5 & 8.9 $\pm$  0.2 & \textcolor{black}{71.3 $\pm$ 0.4} & \textcolor{black}{25.4 $\pm$ 0.7}\\
              IB  & 84.0 $\pm$  0.0 & 27.5 $\pm$  0.0 & 62.1 $\pm$  0.0 & 24.9 $\pm$  0.0 & 65.2 $\pm$  0.0 & 12.8 $\pm$  0.0 & 46.3 $\pm$  0.0 & 6.9 $\pm$  0.0 & \textcolor{black}{84.7 $\pm$ 0.0} & \textcolor{black}{42.7 $\pm$ 0.0}\\
              INVRAT & 87.7 $\pm$ 1.2 & 28.6 $\pm$ 0.9 & 61.8 $\pm$ 1.0 & 30.4 $\pm$ 1.5 & 64.9 $\pm$ 2.5 & 20.8 $\pm$ 1.1 & 47.0 $\pm$ 1.8 & 9.0 $\pm$ 1.5 & \textcolor{black}{83.6 $\pm$ 1.8} & \textcolor{black}{41.4 $\pm$ 1.4}\\
              Inter-RAT & 88.0 $\pm$ 0.7 & 28.9 $\pm$ 0.4 & 62.2 $\pm$ 0.7 & 30.7 $\pm$ 0.5 & 65.8 $\pm$ 0.4 & 21.0 $\pm$ 0.4 & 46.4 $\pm$ 0.6 & 9.9 $\pm$ 0.4 &\textcolor{black}{85.1 $\pm$ 0.5} &\textcolor{black}{43.0 $\pm$ 0.8}\\
              MCD & 89.1 $\pm$ 0.3 &29.1 $\pm$ 0.5 &  62.8 $\pm$ 0.4  & 31.1 $\pm$ 0.6 & 65.2 $\pm$ 0.4 & 23.1 $\pm$ 0.8 & 47.1 $\pm$ 0.6 &10.8 $\pm$ 0.7 &\textcolor{black}{84.4 $\pm$ 0.6} & \textcolor{black}{44.6 $\pm$ 0.2}
               \\
              \midrule
              Vanilla Semi-RAT   & 89.8 $\pm$  0.2  & 30.4 $\pm$  0.2 & 63.3 $\pm$  0.4 & 55.4 $\pm$  0.2 & 57.3 $\pm$  0.3 & 43.0 $\pm$  0.1 & 46.1 $\pm$  0.5 & 25.1 $\pm$  0.2 & \textcolor{black}{82.6 $\pm$ 0.6} & \textcolor{black}{40.7 $\pm$ 0.8}\\
              IB (25\% rationales)  & 85.4 $\pm$  0.0 & 28.2 $\pm$  0.0 & 66.4 $\pm$  0.0 & 54.0 $\pm$  0.0 & 63.4 $\pm$  0.0 & 19.2 $\pm$  0.0 & 46.7 $\pm$  0.0 & 10.8 $\pm$  0.0 & \textcolor{black}{88.8 $\pm$ 0.0} & \textcolor{black}{63.9 $\pm$ 0.0} \\
              WSEE        &90.1 $\pm$  0.1  & 32.2 $\pm$  0.1 & 65.0 $\pm$  0.8 & 55.8 $\pm$  0.5 & 59.9 $\pm$  0.4 & 43.6 $\pm$  0.4 & 49.2 $\pm$  0.9 & 14.8 $\pm$  0.8 & \textcolor{black}{84.3 $\pm$ 0.3} & \textcolor{black}{44.9 $\pm$ 0.5} \\ 
              ST-RAT      & 87.0 $\pm$  0.0 & 31.0 $\pm$  0.0 & - & - & 62.0 $\pm$  0.0 & 51.0 $\pm$  0.0 & 46.0 $\pm$  0.0 & 9.0 $\pm$  0.0 & \textcolor{black}{89.0 $\pm$ 0.0} & \textcolor{black}{\underline{39.0 $\pm$ 0.0}}\\
              \midrule
              Vanilla Sup-RAT  & \underline{93.6 $\pm$  0.3}  & 38.2 $\pm$  0.2 & 63.8 $\pm$  0.2 & 59.4 $\pm$  0.4 & 61.5 $\pm$  0.3 & 51.3 $\pm$  0.2 & 52.3 $\pm$  0.5 & 16.5 $\pm$  0.2 & \textcolor{black}{83.6 $\pm$ 1.4} & \textcolor{black}{68.9 $\pm$ 0.9}\\
              Pipeline       & 86.0 $\pm$  0.0  & 16.2 $\pm$  0.0 & 63.3 $\pm$  0.0 & 41.2 $\pm$  0.0 & \underline{62.3 $\pm$  0.0} & 18.4 $\pm$  0.0 & \underline{70.8 $\pm$  0.0} & \underline{54.8 $\pm$  0.0} & \textcolor{black}{87.7 $\pm$ 0.0} & \textcolor{black}{\underline{81.2 $\pm$ 0.0}}\\
              UNIREX & 91.3 $\pm$ 0.4 & 39.8 $\pm$ 0.6 & 65.5 $\pm$ 0.8 & \underline{62.1 $\pm$ 0.2} & 61.9 $\pm$ 0.7 & 51.4 $\pm$ 0.6 & 48.8 $\pm$ 0.3 & 21.3 $\pm$ 0.1 &\textcolor{black}{81.1 $\pm$ 0.8} &\textcolor{black}{70.9 $\pm$ 0.5} \\
              AT-BMC      & 92.9 $\pm$  0.6 & \underline{40.2 $\pm$  0.3} & \underline{65.8 $\pm$  0.2} & {61.1 $\pm$  0.5} & 62.1 $\pm$  0.2 & \underline{52.1 $\pm$  0.2} & 49.5 $\pm$  0.4 & 18.6 $\pm$  0.3 &  \textcolor{black}{82.3 $\pm$ 0.3} & \textcolor{black}{71.1 $\pm$ 0.6}\\
              \midrule
              $\textmd{SSR}_{unif}$      & 94.3 $\pm$  0.3  & 33.2 $\pm$  0.4 & 62.8 $\pm$  0.3 & 56.2 $\pm$  0.2 & 60.8 $\pm$  0.4 & 47.6 $\pm$  0.5 & 46.8 $\pm$  0.3 & 26.8 $\pm$  0.2 & \textcolor{black}{86.8 $\pm$ 0.9} & \textcolor{black}{46.6 $\pm$ 0.2}\\
              +random DA\;\,             & 90.7 $\pm$  0.3  & 34.5 $\pm$  0.1 & 63.6 $\pm$  0.5 & 56.1 $\pm$  0.3 & {61.3 $\pm$  0.7}  & 48.3 $\pm$  0.5 & 46.0 $\pm$  0.1 & 33.1 $\pm$  0.2 & \textcolor{black}{87.4 $\pm$ 0.3} & \textcolor{black}{47.6 $\pm$ 0.5}\\
              +semantic DA               & 90.7 $\pm$  0.2  & 35.6 $\pm$  0.2 & 64.7 $\pm$  0.7 & 42.7 $\pm$  0.4 & 58.0 $\pm$  0.3 & {50.2 $\pm$  0.3}  & 48.7 $\pm$  0.2 & 33.5 $\pm$  0.4 & \textcolor{black}{87.9 $\pm$ 0.6} & \textcolor{black}{48.0 $\pm$ 0.8}\\
              +mixed DA$\quad$           & {94.5 $\pm$  0.2} & 35.1 $\pm$  0.1 & 65.3 $\pm$  0.6 & 40.3 $\pm$  0.5 & 60.4 $\pm$  0.2 & 49.2 $\pm$  0.5 & 47.6 $\pm$  0.1 & {35.2 $\pm$  0.2}& \textcolor{black}{88.3 $\pm$ 0.3} & \textcolor{black}{48.8 $\pm$ 0.7}\\
              $-$shared $f_{s}$ and $f_{p}$ & 88.3 $\pm$  0.1 & 29.8 $\pm$  0.6 & 60.2 $\pm$  0.3 & 55.7 $\pm$  0.5 & 57.4 $\pm$  0.3 & 43.5 $\pm$  0.2 & 45.6 $\pm$  0.3 & 24.9 $\pm$  0.2 & \textcolor{black}{81.4 $\pm$ 0.3} & \textcolor{black}{39.3 $\pm$ 0.7}\\
              \midrule
              $\textmd{SSR}_{virt}$      & 90.0 $\pm$  0.0  & 34.6 $\pm$  0.2 & 64.2 $\pm$  0.3 & {57.0 $\pm$  0.2}  & 58.2 $\pm$  0.5 & 43.8 $\pm$  0.3 & {50.4 $\pm$  0.3}  & 31.3 $\pm$  0.4 & \textcolor{black}{87.1 $\pm$ 0.4} &\textcolor{black}{47.0 $\pm$ 0.5}\\
              +random DA\;\,             & 92.8 $\pm$  0.2  & 36.7 $\pm$  0.2 & 65.4 $\pm$  0.2 & 44.3 $\pm$  0.4 & 58.3 $\pm$  0.6 & 47.7 $\pm$  0.3 & 46.5 $\pm$  0.3 & 32.4 $\pm$  0.2  & \textcolor{black}{87.4 $\pm$ 0.3} & \textcolor{black}{47.5 $\pm$ 0.9}\\
              +semantic DA               & 87.6 $\pm$  0.3  & 36.9 $\pm$  0.1 & {66.2 $\pm$  0.5}  & 49.8 $\pm$  0.4 & 61.1 $\pm$  0.3 & 48.8 $\pm$  0.2 & 46.5 $\pm$  0.4 & 31.1 $\pm$  0.2 & \textcolor{black}{{88.9 $\pm$ 0.2}} & \textcolor{black}{{49.0 $\pm$ 0.1}}\\
              +mixed DA$\quad$           & 90.5 $\pm$  0.2  & {37.4 $\pm$  0.1}  & 64.5 $\pm$  0.6 & 53.1 $\pm$  0.4 & 60.3 $\pm$  0.2 & 49.1 $\pm$  0.1 & 47.1 $\pm$  0.4 & 33.4 $\pm$  0.3  & \textcolor{black}{88.0 $\pm$ 0.4} & \textcolor{black}{48.5 $\pm$ 0.6} \\
              $-$shared $f_{s}$ and $f_{p}$&  87.9 $\pm$  0.5 & 31.9 $\pm$  0.4 & 62.6 $\pm$  0.3 & 55.3 $\pm$  0.2 & 57.6 $\pm$  0.4 & 43.3 $\pm$  0.5 & 48.6 $\pm$  0.2 & 25.8 $\pm$  0.4 & \textcolor{black}{82.3 $\pm$ 0.5} & \textcolor{black}{40.3 $\pm$ 0.6} \\
              $-$shared $W_{a}$ and $W_{p}$    & 88.3 $\pm$  0.3  & 30.4 $\pm$  0.1 & 62.4 $\pm$  0.9 & 54.0 $\pm$  2.1 & 57.5 $\pm$  0.3 & 42.9 $\pm$  0.1 & 45.8 $\pm$  0.4 & 25.0 $\pm$  0.2 & \textcolor{black}{81.8 $\pm$ 0.7} & \textcolor{black}{39.6 $\pm$ 0.7}\\

              \bottomrule
              \hline
          \end{tabular}
          }
      }   
      \label{table_main}
\end{table*}

\subsubsection{Virtual Shortcuts Representations.}
Intuitively, the supervised rationalization with gold rationales will perform better than the semi-supervised one. However, due to the limited resource, it is difficult for us to annotate the rationales with $\mathcal{D}_{un}$ and further obtain shortcut tokens to improve the performance.
To close the resource gap, we propose a virtual shortcuts representations strategy ($\textmd{SSR}_{virt}$) with transferred shortcuts knowledge from $\mathcal{D}_{sup}$ as guidance.
As shown in Figure \ref{model}, $\textmd{SSR}_{virt}$ also contains two phases (the supervised and unsupervised phase).
Specifically:

\textit{\textbf{In the supervised phase}}, when training the supervised rationalization with $\mathcal{D}_{sup}$, we first 
adopt an \textit{\underline{external predictor}} $q_{\eta}(y|z_{s})$ to predict task results based on the shortcuts $z_{s}$, and ensure the encoder $f_{p_{\eta}}(\cdot)$ in $q_{\eta}$ captures sufficient shortcuts representations $f_{p_{\eta}}(z_{s}) \in \mathbb{R}^{d}$ by minimizing :
\begin{equation}
    \mathcal{L}_{s} = \mathbb{E}_{\substack{z_{s} \sim \mathcal{D}_{sup}}}\left[-\log q_{\eta}(y|z_{s})\right] = \mathbb{E}_{\substack{z_{s} \sim \mathcal{D}_{sup}}}\left[-\log \textrm{softmax}(W_{\eta}f_{p_{\eta}}(z_{s}))\right].
    \vspace{-0.2cm}
\end{equation}

Then, we learn an additional shortcut imitator $f_{a}(x_{sup})$ that takes $x$ in $\mathcal{D}_{sup}$ as the input (denoted by $x_{sup}$ for clarity) to align and mimic $f_{p_{\eta}}(z_{s})$ by minimizing the squared euclidean distance of these two representations, where $f_{a}(\cdot)$ and $f_{p_{\eta}}(\cdot)$ share~parameters:
\begin{equation}
    \mathcal{L}_{virt}=\mathbb{E}_{\substack{ x_{sup},z_{s}  \sim \mathcal{D}_{sup}}}\left[ \left\|f_{p_{\eta}}(z_{s})-f_{a}(x_{sup})\right\|^{2} \right].
    \label{mse}
\end{equation}
\textit{\textbf{In the unsupervised phase}}, during training with $\mathcal{D}_{un}$, we keep $f_{a}(\cdot)$ frozen and employ it to generate virtual shortcuts representations $f_{a}(x_{un})$ by taking $x$ in $\mathcal{D}_{un}$ as the input.
After that, to encourage the model to remove the effect of shortcuts on task predictions, we first adopt $f_{a}(x_{un})$ to match a uniform distribution by calculating 
\begin{equation}
  \mathcal{L}_{diff} =  \mathbb{E}_{\substack{ x \sim \mathcal{D}_{un}}}\left[\operatorname{KL}\left( \mathcal{U}(0, |N|)  \| q_{\sigma}(y|x_{un}) \right)\right],
\end{equation}
where $q_{\sigma}(y|x_{un}) = \textrm{softmax}(W_{a}f_{a}(x_{un}))$.
Next, we set $W_{a}$ and $W_{p}$ share parameters (i.e., $W_{a} \leftrightarrow W_{p}$) to transfer the shortcut information into the \textit{predictor} $f_{p}$, and  further achieve the de-correlation of shortcuts and rationales.
Formally, the final objective of $\textmd{SSR}_{virt}$ is $\mathcal{L}_{{ssr}_{virt}} = \mathcal{L}_{un} + \mathcal{L}_{sup}  + \mathcal{L}_{s} + \lambda_{virt} \mathcal{L}_{virt}+ \lambda_{diff} \mathcal{L}_{diff} $.
\textcolor{black}{Detailed algorithms of $\textmd{SSR}_{virt}$ are shown in Appendix \ref{A.2.algo}.
}


\vspace{-0.2cm}
\subsection{Data Augmentation}
\vspace{-0.2cm}
\label{ssr3.3}
In this section, to close the quantitative gap between $\mathcal{D}_{un}$ and $\mathcal{D}_{sup}$, we propose two data augmentation (DA) methods by utilizing identified shortcuts in $\mathcal{D}_{sup}$.

\textbf{Random Data Augmentation.}
\textcolor{black}{As we have identified the potential shortcuts $z_{s}$ in $\mathcal{D}_{sup}$, we can replace these shortcuts tokens with other tokens which are sampled randomly from the datastore $\mathbb{D}_{random}$. Among them, the database $\mathbb{D}_{random}$ contains all tokens of $\mathcal{D}_{un}$ and $\mathcal{D}_{sup}$ (i.e., $\mathbb{D}_{random}=\left\{x_{j}, \forall x_{j} \in \mathcal{D}_{un} \vee x_{j} \in \mathcal{D}_{sup}\right\}$).}

\begin{table}[!htbp]
  \caption{Task F1 and Token F1 of selected rationales for the four datasets with random DA. }
  \renewcommand\arraystretch{.5}
  \setlength{\tabcolsep}{1.8mm}{
      \scalebox{.78}{
          \begin{tabular}{c|cc|cc|cc|cc}
              \hline    
              \toprule
              \multirow{2}{*}{Methods} &\multicolumn{2}{c|}{Movies} & \multicolumn{2}{c|}{MultiRC} & \multicolumn{2}{c|}{BoolQ} &\multicolumn{2}{c}{Evidence Inference} \\\cmidrule(r){2-9}
                  + random DA&Task&Token-F1&Task&Token-F1&Task&Token-F1 &Task&Token-F1 \\ 
              \midrule

              Vanilla Un-RAT  &88.0 $\pm$ 0.4 &28.4 $\pm$ 0.3 &58.4 $\pm$ 0.2 &24.7 $\pm$ 0.3 &62.1 $\pm$ 0.3 &23.5 $\pm$ 0.2 &47.0 $\pm$ 0.4 &10.4 $\pm$ 0.3 \\

              Vanilla Semi-RAT  &90.6 $\pm$ 0.3 &31.6 $\pm$ 0.1 &64.2 $\pm$ 0.4  &56.2 $\pm$ 0.3  &58.9 $\pm$ 0.1 &44.5 $\pm$ 0.3 &45.0 $\pm$ 0.3  &26.0 $\pm$ 0.3 \\

              WSEE  &89.9 $\pm$ 0.4 &33.4 $\pm$ 0.3  &65.3 $\pm$ 0.1 &55.7 $\pm$ 0.3 &61.0 $\pm$ 0.2 &45.5 $\pm$ 0.3 &50.0 $\pm$ 0.3 &18.7 $\pm$ 0.5  \\
              \textcolor{black}{Vanilla Sup-RAT} & \textcolor{black}{93.0 $\pm$ 0.4} & \textcolor{black}{39.1 $\pm$ 0.3} & \textcolor{black}{64.4 $\pm$ 0.6} & \textcolor{black}{60.6 $\pm$ 0.2} & \textcolor{black}{62.1 $\pm$ 0.4} & \textcolor{black}{51.9 $\pm$ 0.7} & \textcolor{black}{52.8 $\pm$ 0.2} & \textcolor{black}{18.5 $\pm$ 0.3} \\
              
              AT-BMC &92.8  $\pm$  0.1 &40.4 $\pm$ 0.3 &66.6  $\pm$ 0.6 &61.8 $\pm$ 0.5 &62.0 $\pm$ 0.1 &52.6 $\pm$ 0.2 &49.5 $\pm$ 0.3 &19.4  $\pm$ 0.6\\
              \midrule
              $\textmd{SSR}_{unif}$ & 90.7 $\pm$  0.3	&	34.5 $\pm$  0.1	&	63.6 $\pm$  0.5	&	56.1 $\pm$  0.3	&	{61.3 $\pm$  0.7}	&	48.3 $\pm$  0.5	&	46.0 $\pm$  0.1	&	33.1 $\pm$  0.2\\
    
              $\textmd{SSR}_{virt}$   & 92.8 $\pm$  0.2	&	36.7 $\pm$  0.2	&	65.4 $\pm$  0.2	&	44.3 $\pm$  0.4	&	58.3 $\pm$  0.6	&	47.7 $\pm$  0.3	&	46.5 $\pm$  0.3	&	32.4 $\pm$  0.2\\
                  \bottomrule
              \hline

          \end{tabular}
          }
}   
      \label{table_da}
    \end{table}

\textbf{Semantic Data Augmentation.}
Besides the random augmentation, we design a retrieval-grounded semantic augmentation method by replacing shortcut tokens with several tokens semantically close to them through retrieval.
\textcolor{black}{Detailed retrieval algorithms about semantic DA are shown in Appendix \ref{app_algo_da}. 
Besides, we also mix data augmented from random DA with data augmented from semantic DA to achieve mixed data augmentation.}

\section{Experiments}
\subsection{Datasets and Comparison Methods}
\label{4.1}

\textbf{Datasets.}
We evaluate SSR on text classification tasks from the ERASER benchmark \citep{DeYoungJRLXSW20}, including Movies \citep{pang2004sentimental} for sentiment analysis, MultiRC \citep{khashabi2018looking} for multiple-choice QA, BoolQ \citep{clark-etal-2019-boolq} for reading comprehension, Evidence Inference \citep{lehman-etal-2019-inferring} for medical interventions, and FEVER \citep{thorne2018fever} for fact verification. 
Each dataset contains human annotated rationales and classification labels. In the semi-supervised (or unsupervised) setting, partially (or fully) labeled rationales are unavailable.


\textbf{Comparison Methods.}
We~compare~SSR~against~three~type~methods~as~follows:

\noindent$\bullet$ \textbf{Unsupervised rationalization:}
\textbf{Vanilla Un-RAT} is the method presented in section \ref{2.2}, which samples rationale tokens from a Bernoulli distribution of each token with a Gumbel-softmax reparameterization. 
In practice, we employ Vanilla Un-RAT as the unsupervised rationalization method $\mathcal{M}_{un}$ in section \ref{3.1}.
\textbf{IB} \citep{paranjape-etal-2020-information} employs an Information Bottleneck \citep{AlemiFD017} principle to manage the trade-off between achieving accurate classification performance and yielding short rationales.
\textbf{INVRAT} \citep{chang2020invariant} learns invariant rationales by exploiting multiple environments to remove shortcuts in data.
\textbf{Inter-RAT} \citep{yue2022interventional} proposes a causal intervention method to remove spurious correlations in selective rationalization.
\textbf{MCD} \citep{shabi}  uncover the conditional
independence relationship between the target label and non-causal and causal features to compose rationales.



\noindent$\bullet$ \textbf{Supervised rationalization:}
\textbf{Vanilla Sup-RAT} is the method we describe in section \ref{2.3}, which trains task classification and rationalization jointly.
\textbf{Pipeline} \citep{lehman-etal-2019-inferring} is a pipeline model which trains the \textit{selector} with gold rationales and \textit{predictor} with class labels independently.
\textbf{UNIREX} \citep{chan2022unirex} proposes a unified supervised rationalization framework to compose faithful and plausible rationales.
\textbf{AT-BMC} \citep{li2022unifying} is the state-of-the-art (SOTA) supervised rationalization approach, which is implemented with label embedding and mixed  adversarial training.



\noindent$\bullet$ \textbf{Semi-supervised rationalization:}
\textbf{Vanilla Semi-RAT} is the method described in section \ref{2.4}, which can be seen as the ablation of SSR (i.e., without exploiting shortcuts).
\textbf{IB (25\% rationales)} \citep{paranjape-etal-2020-information} trains the \textit{selector} with 25\% annotated rationales through the BCE loss, and employs the rest data to train the unsupervised IB.
\textbf{WSEE} \citep{pruthi-etal-2020-weakly} proposes a classify-then-extract framework, conditioning rationales extraction on the predicted label.
\textbf{ST-RAT}~\citep{bhat2021self} presents a self-training framework by exploiting the pseudo-labeled rationale examples.





\begin{figure}[h]
  \begin{minipage}[t]{0.4\linewidth}
    \centering
    \includegraphics[width=4.5cm]{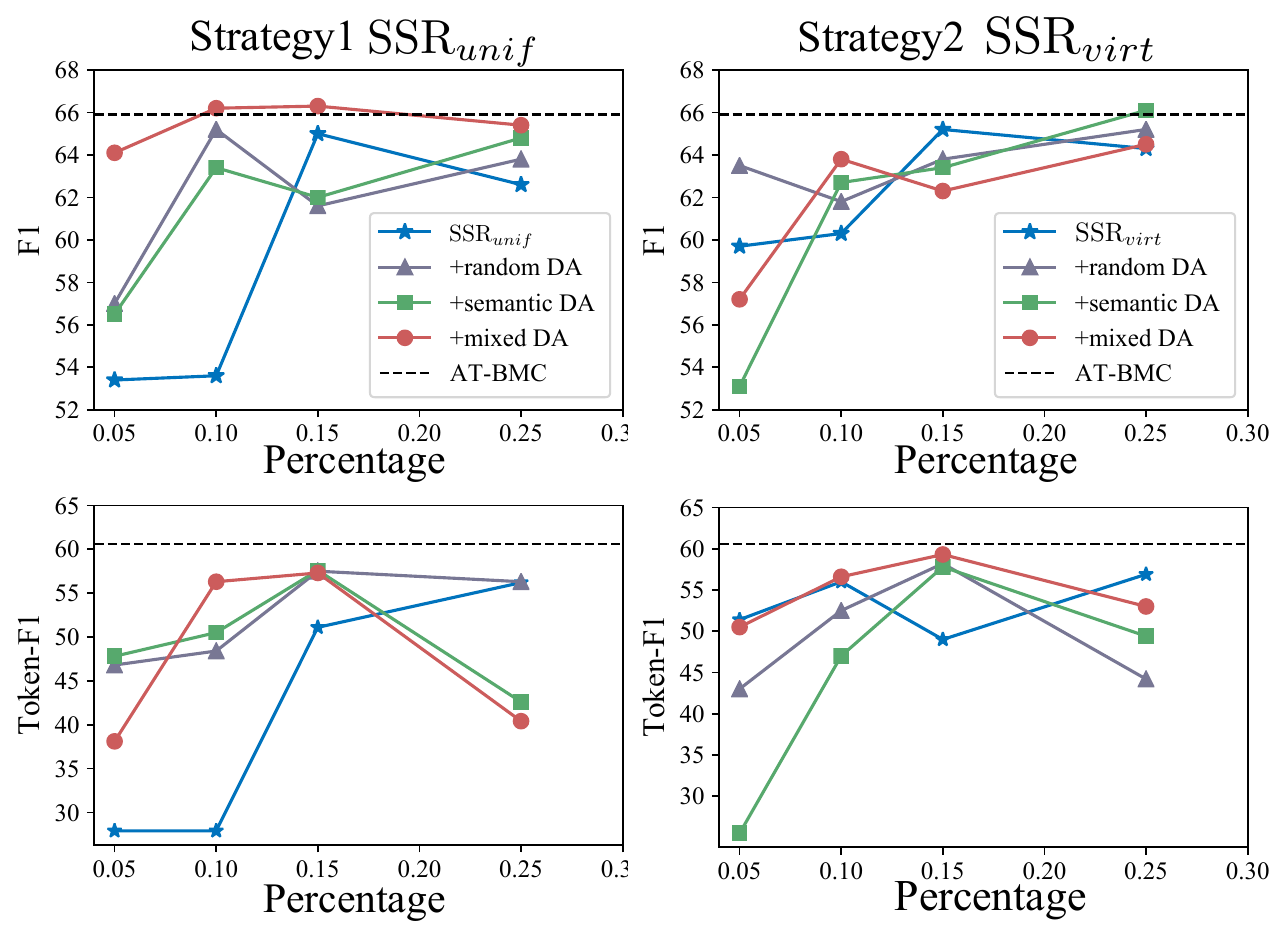}
    \caption{Gold Rationale Efficiency.
    }
    \label{gold}
\end{minipage}
\hfill
  \begin{minipage}[t]{0.6\linewidth}
      \centering
      \includegraphics[width=7.6cm]{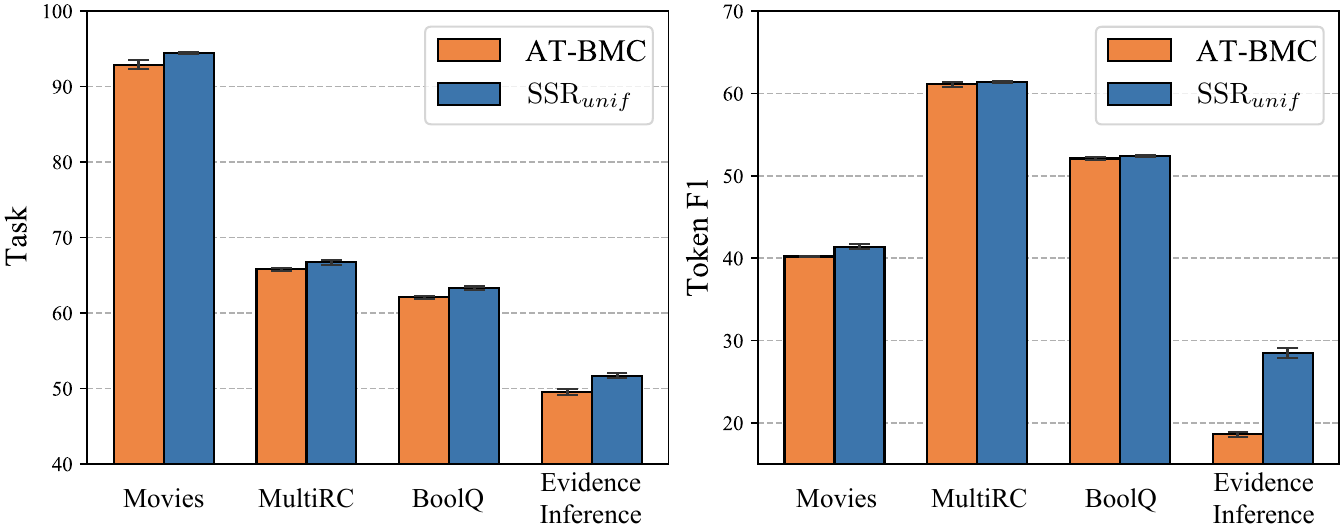}
      \caption{$\textmd{SSR}_{unif}$ with full annotations.}
      \label{full_an}
  \end{minipage}
 \vspace{-0.4cm}
\end{figure}

\subsection{Experimental Setup}
Following prior researches \citep{paranjape-etal-2020-information,bhat2021self}, we employ BERT \citep{devlin-etal-2019-bert} as the encoder in both the \textit{selector}  and \textit{predictor}.
For training, we adopt  the AdamW optimizer \citep{LoshchilovH19} with an initial learning rate as 2e-05, then we set the batch size as 4, maximum sequence length as 512 and training epoch as 30.
Besides, we set the predefined sparsity level $\alpha$ as $\{0.1,0.2,0.2,0.08,0.2\}$ for Movies, MultiRC, BoolQ, Evidence Inference and FEVER, respectively, which is slightly higher than the percentage of rationales in the input text. 
In the semi-supervised setting, we implement our SSR and other semi-supervised rationalization methods with 25\% labeled rationales.
In SSR, we set $\mathcal{L}_{unif}$, $\mathcal{L}_{virt}$, and $\lambda_{diff}$ as 0.1, respectively.
For evaluation, we report weighted F1 scores for classification accuracy following \citep{paranjape-etal-2020-information,bhat2021self}.

Then, to evaluate quality of rationales, we report token-level F1 scores.
For a fair comparison, results of IB, Pipeline and ST-RAT in Table \ref{table_main} are directly taken  \citep{paranjape-etal-2020-information,bhat2021self}.
Besides, as WSEE reports Macro F1 and AT-BMC reports Micro F1 for task classification, we re-run their released codes  \citep{pruthi-etal-2020-weakly,li2022unifying} and present weighted F1 scores.


\subsection{Experimental Results}
\vspace{-0.1cm}
\textbf{Overall Performance.} 
We compare SSR with baselines across all the datasets, and experimental results are shown in Table \ref{table_main}. From the results, in general, we observe supervised rationalization methods perform the best, followed by semi-supervised rationalization, and unsupervised methods are the worst.
Compared with the semi-supervised rationalization (e.g., ST-RAT), both $\textmd{SSR}_{unif}$ and $\textmd{SSR}_{virt}$ achieve promising performance, indicating the effectiveness of exploiting shortcuts to compose rationales.
Besides, after data augmentations, our approach achieves a similar performance to the SOTA supervised rationalization method (i.e., AT-BMC).
Considering that our method adopts only 25\% annotated rationales, we argue that it is sufficient to demonstrate the strength of replacing the shortcut tokens for data augmentations.
Finally, we compare $\textmd{SSR}_{unif}$ and $\textmd{SSR}_{virt}$. 
From the observation, we conclude $\textmd{SSR}_{virt}$ performs better than $\textmd{SSR}_{unif}$ in most datasets, illustrating employing virtual shortcuts representations to rationalization may be more effective with few labeled~rationales.

\textbf{Ablation study.}  For $\textmd{SSR}_{unif}$, we first let encoder $f_{s}$ and $f_{p}$ not share parameters and the results ($-$shared $f_{s}$ and $f_{p}$) are shown in the Table \ref{table_main}. We can find the results degrade significantly after removing the shared linear parameters. Then, we remove the uniform constraint. Since we implement Vanilla Sup-RAT with the same set of shared parameters, we argue that Vanilla Sup-RAT can be seen as the ablative variant of  which removes the uniform constraint. From Table \ref{table_main} in the paper, we find $\textmd{SSR}_{unif}$ outperforms Vanilla Sup-RAT.
For $\textmd{SSR}_{virt}$, we set the encoder $f_{s}$ and $f_{p}$ not share parameters, $W_{a}$ and $W_{p}$ not share parameters, respectively, and the results are shown in the Table~\ref{table_main}. We also find the results degrade significantly after removing the shared encoder or the shared $W_{a}$ and $W_{p}$.
Besides, the results of without the shared $W_{a}$ and $W_{p}$ perform worse than without shared encoders $f_{s}$ and $f_{p}$, indicating the effectiveness of the shared  $W_{a}$ and $W_{p}$.
From the observations, we can conclude that the components of $\textmd{SSR}_{unif}$ and $\textmd{SSR}_{virt}$ are necessary. 





   


\textbf{Analysis on Data Augmentation.}
We develop two data augmentation methods, including random DA and semantic DA. 
\textcolor{black}{We augment the data with 25\% of the original dataset for both random DA and semantic DA.}
From Table \ref{table_main}, we find SSR with semantic DA performs better than random DA, 
which validates that replacing shortcut tokens with semantically related tokens is more effective. 
Interestingly, SSR with mixed DA does not always perform better than semantic DA, and we argue a potential reason is that since the augmented data only replace part of the input tokens and most tokens in the text remain unchanged, the augmented data have many tokens that are duplicated from the original data, and too much of such data does not be beneficial for model training and even degrade~it.
Besides, we compare with baselines with the same data augmentation methods. Table \ref{table_da} shows all baselines implemented with the random DA.
With the same amount of data, our SSR achieves competitive results, especially in Token F1.
\textcolor{black}{More experimental results are shown in Appendix \ref{appda}.}

\textbf{Gold Rationale Efficiency.}
After the analysis on DA, we also find the performance of SSR with DA degrades a lot on the MultiRC dataset.
Therefore, in this section, we investigate SSR performance with varying proportions of annotated rationales in the training set to see how much labeled data is beneficial for MultiRC.
We make experiments on the MultiRC dataset and report corresponding results in Figure \ref{gold}.
From the figure, we observe both F1 and Token F1 of SSR increase with increasing proportions until 15\%, and degrade when proportions exceed 15\%. Meanwhile, the performance between $\textmd{SSR}_{unif}$+mixed DA with 15\% gold rationales and AT-BMC is not significant.
The above observation illustrates that more labeled rationales may be not better and SSR can effectively compose rationales and yield results without extensive manual rationales by exploring~shortcuts. \textcolor{black}{More experiments can be found in Appendix \ref{c.1.new}.}



\textbf{SSR with Full Annotations.}
We investigate the performance of SSR with full annotations (i.e., the proportions of annotated rationales achieve 100\%.).
Specifically, for strategy 2, since there exist no shortcuts need to be ``virtual'', $\textmd{SSR}_{virt}$ is unavailable.
For strategy 1, $\textmd{SSR}_{unif}$  with full annotations means it has incorporated all identified shortcuts in the supervised phase.
In Figure \ref{full_an}, we compare $\textmd{SSR}_{virt}$ with full annotations with AT-BMC that also exploits all labeled rationales. From the figure, we find $\textmd{SSR}_{virt}$  outperforms AT-BMC, indicating introducing all shortcuts into SSR is effective.

Besides, we also implement AT-BMC on Movies by~introducing shortcuts (i.e., replace the original objective of AT-BMC $\mathcal{L}_{at-bmc}$ as $\mathcal{L}_{at-bmc} + \mathcal{L}_{unif}$ with Eq (\ref{kl})).
The corresponding F1 and Token F1 scores are \textbf{\underline{94.7}} and \textbf{\underline{43.2}}, which still perform better than the original AT-BMC.
Such observations strongly demonstrate that we can boost supervised rationalization by introducing shortcuts~explicitly.

\begin{wraptable}[11]{r}{5.35cm}
   \vspace{-.6cm}
 \setlength{\belowcaptionskip}{0.2cm}
 \caption{Generalization Evaluation on IMDB and SST-2.}
 \vspace{-0.3cm}
 \centering
 \renewcommand\arraystretch{0.8}
 \setlength{\tabcolsep}{2.8mm}{
 \scalebox{0.75}{
   \begin{tabular}{ccc}
     \toprule
             Methods   & IMDB & SST-2\\
     \midrule
         Vanilla Un-RAT                       & 85.3 $\pm$ 0.2 & 45.3 $\pm$ 8.1   \\
         +semantic DA  & 86.6 $\pm$ 0.4 & 47.8 $\pm$ 6.6 \\
         Vanilla Semi-RAT                 & 89.5 $\pm$ 0.4  & 75.9 $\pm$ 0.7\\
         +semantic DA	&89.7	$\pm$ 0.3 &76.4	$\pm$ 0.5 \\
         WSEE &90.5 $\pm$ 0.3	&77.1	 $\pm$ 0.6\\
         +semantic DA	&\textbf{91.0	$\pm$ 0.4} &78.3 $\pm$	 0.7\\
         \midrule
         $\textmd{SSR}_{unif}$    & 90.3 $\pm$  0.2&  79.4 $\pm$ 0.3  \\
         +semantic DA             & 90.7 $\pm$ 0.1& 82.4 $\pm$ 0.8\\
 
         $\textmd{SSR}_{virt}$    & 89.9 $\pm$ 0.2 & 79.9 $\pm$ 0.4\\
         +semantic DA             & 90.3 $\pm$ 0.3 & \textbf{83.5  $\pm$ 0.5  }\\
         
     \bottomrule
     \label{rationales}
 \end{tabular}
    }
 }   
 \label{table_movie}
 \end{wraptable}
\textbf{Generalization Evaluation.}
Since SSR explicitly removes the effect of shortcuts on yielding task results and composing rationales, SSR can generalize better to out-of-distribution (OOD) datasets than unsupervised rationalization methods, where such unsupervised methods generalize poorly since the shortcuts are changed.
To this end, we conduct an experiment to validate this opinion.
Specifically, we introduce a new movie reviews dataset SST-2 \citep{socher2013recursive} which contains pithy expert movie reviews.
As the original Movies dataset in section \ref{4.1} contains lay movie reviews \citep{hendrycks-etal-2020-pretrained}, SST-2 can be considered as the OOD dataset corresponding to Movies.
Meanwhile, we make experiments on an identically distributed dataset IMDB \citep{maas2011learning}.
Since there exist no labeled rationales in IMDB and SST-2, we investigate the model performance by calculating weighted F1 scores. In Table \ref{rationales}, we find all models achieve promising results on IMDB.
However, when evaluating on SST-2,  F1 scores of SSR are much higher than baselines, indicating the effectiveness of exploring shortcuts to predict~task~results.

\textbf{Visualizations.}
We provide a qualitative analysis on rationales extracted by SSR in Appendix~\ref{appvis}.
By showing several examples of rationales selected by Vanilla Un-RAT and our $\textmd{SSR}_{unif}$, we conclude that $\textmd{SSR}_{unif}$ can avoid to  extract shortcuts effectively. \textcolor{black}{Besides, we make some subjective evaluations to evaluate extracted rationales in Appendix \ref{appchat}}, where we find $\textmd{SSR}_{unif}$ outperforms baselines in all subjective metrics (e.g. usefulness and completeness), illustrating the effectiveness of $\textmd{SSR}_{unif}$.

\section{Conclusions}
In this paper, we proposed a Shortcuts-fused Selective Rationalization (SSR) method, improving rationalization by incorporating shortcuts explicitly.
To be specific, we first developed a shortcut discovery approach to obtain several potential shortcut tokens.
Then, we designed two strategies to augment the identified shortcuts into rationalization, mitigating the problem of employing shortcuts to compose rationales and yield classification results.
Finally, we further utilized shortcuts for data augmentation by replacing shortcut tokens with random or semantic-related tokens.
Experimental results on real-world datasets clearly demonstrated the effectiveness of our proposed method.

\textbf{Acknowledgements.}
This research was supported by grants from the National Key Research and Development Program of China (Grant No. 2021YFF0901003), the National Natural Science Foundation of China (Grants No. 62337001, 62337001), and the Fundamental Research Funds for the Central Universities.
\newpage


\bibliography{iclr2024_conference}
\bibliographystyle{iclr2024_conference}
\clearpage
\appendix
\section{Algorithms}
\subsection{\textcolor{black}{Algorithm of $\textmd{SSR}_{unif}$}}
\label{A.1.algo}
\begin{algorithm}[H]
  \setstretch{1.} 
  \caption{$\textmd{SSR}_{unif}$: Injecting Shortcuts into Prediction.} 
  \begin{algorithmic}
    \State \textit{\textbf{In the supervised phase:}}
    \State 1. Calculate the original  supervised rationalization loss $\mathcal{L}_{sup}$.
    \State 2. Ensure the \textit{predictor} $q_{\psi}$ identifies the shortcuts $z_{s}$ as meaningless features:
    \State \quad $\mathcal{L}_{unif}=\mathbb{E}_{\substack{ z_{s}  \sim \mathcal{D}_{sup}}}\left[\operatorname{KL}\left( \mathcal{U}(0, |N|) \|  q_{\psi}(y|z_{s})\right)\right]$.
    \State \textit{\textbf{In the unsupervised phase:}}
    \State 3. Calculate the original unsupervised rationalization loss $\mathcal{L}_{un}$.
    \State 4. Set $W_{p_{un}}f_{p_{un}}(\cdot)$ and $W_{p_{sup}}f_{p_{sup}}(\cdot)$ sharing parameters:
    \State \quad $W_{p_{un}}f_{p_{un}}(\cdot)\leftrightarrow W_{p_{sup}}f_{p_{sup}}(\cdot)$
    \State 5. Calculate the objective of $\textmd{SSR}_{unif}$:
    \State \quad $\mathcal{L}_{ssr_{unif}} = \mathcal{L}_{un} + \mathcal{L}_{sup} + \lambda_{unif} \mathcal{L}_{unif}$.

  \end{algorithmic}
  \label{algorithm1}
\end{algorithm}

\subsection{\textcolor{black}{Algorithm of $\textmd{SSR}_{virt}$}}
\label{A.2.algo}
\begin{algorithm}[H]
  \setstretch{1.} 
  \caption{$\textmd{SSR}_{virt}$: Virtual Shortcuts Representations.} 
  \begin{algorithmic}
    \State \textit{\textbf{In the supervised phase:}}
    \State 1. Calculate the original  supervised rationalization loss $\mathcal{L}_{sup}$.
    \State 2. Employ an \textit{\underline{external predictor}} $q_{\eta}(y|z_{s})$ to predict task results based on the shortcuts $z_{s}$:
    \State \quad  $\mathcal{L}_{s} = \mathbb{E}_{\substack{z_{s} \sim \mathcal{D}_{sup}}}\left[-\log q_{\eta}(y|z_{s})\right] = \mathbb{E}_{\substack{z_{s} \sim \mathcal{D}_{sup}}}\left[-\log \textrm{softmax}(W_{\eta}f_{p_{\eta}}(z_{s}))\right]$.
    \State 3. Learn an additional shortcut imitator $f_{a}(x_{sup})$ that takes $x_{sup}$ in $\mathcal{D}_{sup}$ to learn the shortcut representation $f_{p_{\eta}}(z_{s})$:
    \State \quad $\mathcal{L}_{virt}=\mathbb{E}_{\substack{ x_{sup},z_{s}  \sim \mathcal{D}_{sup}}}\left[ \left\|f_{p_{\eta}}(z_{s})-f_{a}(x_{sup})\right\|^{2} \right]$.

    \State \textit{\textbf{In the unsupervised phase:}}
    \State 4. Calculate the original unsupervised rationalization loss $\mathcal{L}_{un}$.
    \State 5. Generate virtual shortcuts representations $f_{a}(x_{un})$ by taking $x_{un}$ in $\mathcal{D}_{un}$.
    \State 6. Employ $f_{a}(x_{un})$ to match a uniform distribution by calculating:
    \State \quad  $\mathcal{L}_{diff} =  \mathbb{E}_{\substack{ x \sim \mathcal{D}_{un}}}\left[\operatorname{KL}\left( \mathcal{U}(0, |N|)  \| q_{\sigma}(y|x_{un}) \right)\right]$.
    \State 7. Set $W_{a}$ and $W_{p}$ share parameters: $W_{a}\leftrightarrow W_{p}$
    \State 8. Calculate the objective of $\textmd{SSR}_{virt}$:
    \State \quad $\mathcal{L}_{ssr_{virt}} = \mathcal{L}_{un} + \mathcal{L}_{sup}  + \mathcal{L}_{s} + \lambda_{virt} \mathcal{L}_{virt}+ \lambda_{diff} \mathcal{L}_{diff}$.

  \end{algorithmic}
  \label{algorithm2}
\end{algorithm}

\subsection{Algorithm of Semantic Data Augmentation}
\label{app_algo_da}
Besides the random augmentation, we design a retrieval-grounded semantic augmentation method by replacing shortcut tokens with several tokens semantically close to them through retrieval.
To achieve this goal, we first construct a global datastore $\mathbb{D}_{global}$ consisting of a set of key-value pairs offline, where the key is a $d$-dimensional representation of the input computed by the encoder $f_{p_{un}}(\cdot)$ in $\mathcal{M}_{un}$ (section~\ref{3.1}) and the value is  the corresponding input.
Then, for each input $x$ in $\mathcal{D}_{sup}$, we can search another input $x^{r}$ that is nearest to its semantics (except for itself) by retrieving this~datastore. In this paper, we compute the L2 distance between the two input representations to indicate the semantic relevance, where the smaller the L2 distance, the closer an input semantically to another.

Next, we build a local datastore $\mathbb{D}_{local}$ by employing $f_{p_{un}}(\cdot)$ to represent each token $x^{r}_{i}$ in $x^{r}$, where the value is $x^{r}_{i}$ and key is $f_{p_{un}}(x^{r}_{i})$.
Then, we adopt $f_{p_{un}}(x_{i})$ to retrieve the nearest token in $x^{r}$ to $x_{i}$ with the $\mathbb{D}_{local}$.
In the retrieval process, we first calculate the L2 distance between the semantic representation of the retrieved word $x_i$ and each word $x_{i}^{r}$ in the database. Next, we choose the word with the closest L2 distance to the retrieved word $x_i$ as the semantically similar word. In the specific code implementation, we employ F{\small AISS} \citep{johnson2019billion} to achieve this retrieval goal.

\begin{algorithm}[H]
  \setstretch{1.} 
  \caption{Semantic Data Augmentation} 
  \begin{algorithmic}

    \State \textbf{Input:} Supervised dataset $\mathcal{D}_{sup}$, and a well-trained encoder $f_{p_{un}}(\cdot)$ in unsupervised rationalization model $\mathcal{M}_{un}$.
    \State \textbf{Output:} Several semantic related tokens.
    \State \textbf{Create a global datastore $\mathbb{D}_{global}$:}
    \For{j=1 \textbf{to} $|\mathcal{D}_{sup}|$}
    \State Sample $x$ from $\mathcal{D}_{sup}$.
    \State Construct a key-value pair: 
     (key, value) = $\left(f_{p_{un}}(x), x \right)$
    \EndFor
    \State  $\mathbb{D}_{global}=\left\{\left(f_{p_{un}}(x), x \right), \forall x \in \mathcal{D}_{sup}\right\}$.
    \State Search the nearest semantic $x^{r}$ to $x$ in $\mathbb{D}_{global}$, $x^{r} \neq x$.
    \State \textbf{Create a local datastore $\mathcal{D}_{local}$:}
    \For{i=1 \textbf{to} $|x^{r}|$}
    \State (key, value) = $\left(f_{p_{un}}(x_{i}^{r}), x_{i}^{r} \right)$.
    \EndFor
    \State  $\mathbb{D}_{local}=\left\{\left(f_{p_{un}}(x_{i}^{r}), x_{i}^{r} \right), \forall x_{i}^{r} \in x^{r}\right\}$.
    \State Search the nearest semantic token $x_{i}^{r}$ to $x_{i}$ in $\mathbb{D}_{local}$, $x_{i}^{r}$ does not belong to gold tokens in $x$ or $x^{r}$.

  \end{algorithmic}
  \label{algorithm}
\end{algorithm}

Finally, we replace shortcuts tokens with retrieved tokens to achieve semantic augmentation.
It is worth noting that the goal of the semantic augmentation is to retrieve several tokens that are semantically similar to shortcut tokens. However, it is still possible to retrieve gold tokens. 
To avoid this, in our implementation, when the retrieved token belongs to gold tokens in $x$ or $x^{r}$, we will filter it and search the next token. Detailed algorithms about semantic DA are present in Algorithm \ref{algorithm}.

\section{Related Work}
Selective Rationalization has achieved significantly progress recently.
Existing approaches~\citep{lei-etal-2016-rationalizing,bao-etal-2018-deriving,antognini2021multi,AntogniniF21,lakhotia-etal-2021-fid,CartonKT22,shabi} can be categorized into~three types, including unsupervised, supervised and semi-supervised rationalization.
In unsupervised rationalization, \cite{lei-etal-2016-rationalizing} first proposed a classical framework consisting of a \textit{selector} and \textit{predictor}. Following this framework, \cite{bastings-etal-2019-interpretable}
studied a HardKuma reparameterization to replace REINFORCE in \citep{lei-etal-2016-rationalizing}. \cite{paranjape-etal-2020-information,ChenJ20} balanced the task accuracy and sparsity of rationales with an information bottleneck regular. \cite{chang2020invariant} discovered the causal and invariant rationales by creating different environments.
Based~on the \textit{selector}-\textit{predictor} framework, several approaches 
expanded it by introducing an external \textit{guider}.
Among them, \cite{CDY21,sha2021learning} considered the original rationalization method as the \textit{student}, and a well-trained network as the \textit{teacher}. Then, they adopted the \textit{teacher} to guide the \textit{student} to select rationales.
Different from them, \cite{yuedare} developed a ``self-guided'' pattern, which explored non-rationale tokens to guide the rationale generation with a disentangled method.

In the supervised rationalization, \cite{DeYoungJRLXSW20} studied an ERASER benchmark which contains several datasets with both task labels and gold rationales.
In ERASER, a pipeline approach \citep{lehman-etal-2019-inferring} was proposed to be a classical baseline in supervised rationalization.
\cite{chan2022unirex} developed a unified framework to train classification and rationalization jointly.
\cite{li2022unifying} proposed to employ mixed adversarial training and boundary
match constraint to improve rationalization (the SOTA model in supervised~rationalization).

Since annotating rationales is time-consuming and labor-intensive, several researches \citep{paranjape-etal-2020-information,pruthi-etal-2020-weakly,bhat2021self} focused on the semi-supervised rationalization.
Among them, \cite{paranjape-etal-2020-information} first experimented with a semi-supervised setting.
\cite{bhat2021self} developed a multi-task teacher-student framework based on a self-training pattern. They employed gold rationales to train a supervised rationalization method, and adopted it to label the unsupervised data to obtain pseudo-labeled examples to boost semi-supervised rationalization.

\section{Setting}
\subsection{\textcolor{black}{Gumbel-Softmax in Rationalization.}}
\label{gumbel}
Gumbel-Softmax is a commonly used technique for handling discrete data in generative models and optimization problems. It combines the Gumbel distribution and the Softmax function to sample from a discrete probability distribution. The key idea is to introduce noise from the Gumbel distribution and then transform this noise into a sample from a discrete distribution using the Softmax function.

The process of Gumbel-Softmax can be summarized as follows:

Sampling noise from the Gumbel distribution: First, a noise vector $g$ is sampled from the Gumbel(0, 1) distribution, where $0$ is the location parameter and $1$ is the scale parameter. This noise vector $g$ introduces randomness into the sampling process.

Computing the Gumbel-Softmax sample: Next, the noise vector $g$ is added to the logarithmic values of the discrete probability distribution $p$. Then, the Softmax function is applied to obtain a sample from the discrete distribution. Specifically, for the logarithmic value $z_i$ of the discrete probability distribution, the Gumbel-Softmax sample is calculated as follows:

\begin{equation}
  \text{Gumbel-Softmax}(z_i) = \frac{\exp((\log(p_i) + g_i)/\tau)}{\sum_{j=1}^{K} \exp((\log(p_j) + g_j)/\tau)},
\end{equation}

where $\tau$ is the temperature parameter that controls the smoothness of the sample. Higher temperature values result in smoother samples, while lower temperature values tend to produce more one-hot vectors that represent discrete values. By adjusting the temperature parameter $\tau$, the randomness and smoothness of the Gumbel-Softmax samples can be controlled. As $\tau$ approaches $0$, the sample tends to be a one-hot vector, where only one element is $1$ and the rest are $0$, resembling maximum likelihood estimation. As $\tau$ approaches positive infinity, the sample tends to approach a uniform distribution, where all elements have equal probabilities.

The benefit of Gumbel-Softmax is that it provides a differentiable approximation for sampling discrete variables, making it compatible with optimization algorithms such as gradient descent. This property makes it applicable in unsupervised rationalization to sample rationale tokens.

\subsection{Shared Parameters}
\label{app_share}
Table \ref{share-p} lists all shared parameters (each row's parameters are shared):
\begin{table}[!htbp]
   \caption{Detailed shared parameters.}
   \vspace{-0.3cm}
   \renewcommand\thetable{2}
   \renewcommand\arraystretch{2.}
   \centering
   \renewcommand\arraystretch{1.2}
   \setlength{\tabcolsep}{10.3mm}{
       \scalebox{.9}{
   \begin{tabular}{cc}
   \toprule
   Supervised Phase   & Unsupervised Phase \\
   \midrule
   $W_{s_{sup}}$ in the \textit{selector}  & $W_{s_{un}}$ in the \textit{selector}  \\
   $W_{p_{sup}}$ in the \textit{predictor} & $W_{p_{un}}$ in the \textit{predictor} \\
   $f_{s_{sup}}(\cdot)$           & $f_{s_{un}}(\cdot)$           \\
   $f_{p_{sup}}(\cdot)$           & $f_{p_{un}}(\cdot)$           \\
   $f_{s_{sup}}(\cdot)$           & $f_{p_{un}}(\cdot)$           \\
   $f_{p_{sup}}(\cdot)$           & $f_{s_{un}}(\cdot)$           \\
       
   \bottomrule
   \label{share-p}
 \end{tabular}
       }}
 \end{table}

\subsection{Missed Results}
In this paper, we do not report ST-RAT results on MultiRC data.
Although ST-RAT has reported its code URL \url{https://aka.ms/RationaleST} in the paper, we still fail to find the corresponding codes. Therefore,  it is difficult for us to reproduce the  results of ST-RAT. Consequently,  we directly use the results in the original ST-RAT paper. Since ST-RAT is not experimented on MultiRC, we do not report the result of ST-RAT on MultiRC.



\section{More Experimental Results}
\subsection{More Baselines}
In this section, we add two unsupervised rationalization methods as our baselines, including CAR \citep{chang2019game} and {3Player} \citep{yu2019rethinking}:

$\bullet$
\textcolor{black}{\textbf{CAR} \citep{chang2019game} proposes a game theoretic approach to  rationalization.}

$\bullet$
\textcolor{black}{\textbf{3Player} \citep{yu2019rethinking} adopts an introspective model which  predicts and incorporates the outcome into the rationalization.}

The overall experimental results are shown in Table \ref{table_main2-app}.

\begin{table*}[htbp]
  \caption{Task F1 and Token F1 of selected rationales for the five datasets. Among them, the underlined scores are the state-of-the-art performances of the supervised rationalization. }
  \renewcommand\arraystretch{.4}
  \setlength{\tabcolsep}{2.73mm}{
      \scalebox{.58}{
          \begin{tabular}{c|cc|cc|cc|cc|cc}
              \hline    
              \toprule
                  \multirow{2}{*}{Methods} &\multicolumn{2}{c|}{Movies} & \multicolumn{2}{c|}{MultiRC} & \multicolumn{2}{c|}{BoolQ} &\multicolumn{2}{c}{Evidence Inference} & \multicolumn{2}{c}{FEVER}\\\cmidrule(r){2-11}
                  &Task&Token-F1&Task&Token-F1&Task&Token-F1 &Task&Token-F1&Task&Token-F1 \\ 
              \midrule
              Vanilla Un-RAT  &87.0 $\pm$  0.1	&	28.1 $\pm$  0.2	&	57.7 $\pm$  0.4	&	23.9 $\pm$  0.5	&	62.0 $\pm$  0.2	&	19.7 $\pm$  0.4	&	46.2 $\pm$  0.5	&	8.9 $\pm$  0.2 & \textcolor{black}{71.3 $\pm$ 0.4} & \textcolor{black}{25.4 $\pm$ 0.7}\\
              IB  & 84.0 $\pm$  0.0	&	27.5 $\pm$  0.0	&	62.1 $\pm$  0.0	&	24.9 $\pm$  0.0	&	65.2 $\pm$  0.0	&	12.8 $\pm$  0.0	&	46.3 $\pm$  0.0	&	6.9 $\pm$  0.0 & \textcolor{black}{84.7 $\pm$ 0.0} & \textcolor{black}{42.7 $\pm$ 0.0}\\
              INVRAT & 87.7 $\pm$ 1.2 & 28.6 $\pm$ 0.9 & 61.8 $\pm$ 1.0 & 30.4 $\pm$ 1.5 & 64.9 $\pm$ 2.5 & 20.8 $\pm$ 1.1 & 47.0 $\pm$ 1.8 & 9.0 $\pm$ 1.5 & \textcolor{black}{83.6 $\pm$ 1.8} & \textcolor{black}{41.4 $\pm$ 1.4}\\
              Inter-RAT & 88.0 $\pm$ 0.7 & 28.9 $\pm$ 0.4 & 62.2 $\pm$ 0.7 & 30.7 $\pm$ 0.5 & 65.8 $\pm$ 0.4 & 21.0 $\pm$ 0.4 & 46.4 $\pm$ 0.6 & 9.9 $\pm$ 0.4 &\textcolor{black}{85.1 $\pm$ 0.5} &\textcolor{black}{43.0 $\pm$ 0.8}\\
              MCD & 89.1 $\pm$ 0.3 &29.1 $\pm$ 0.5 &  62.8 $\pm$ 0.4  & 31.1 $\pm$ 0.6 & 65.2 $\pm$ 0.4 & 23.1 $\pm$ 0.8 & 47.1 $\pm$ 0.6 &10.8 $\pm$ 0.7 &\textcolor{black}{84.4 $\pm$ 0.6} & \textcolor{black}{44.6 $\pm$ 0.2}
               \\
               CAR     & \textcolor{black}{84.8 $\pm$ 0.2} & \textcolor{black}{28.3 $\pm$ 0.5} & \textcolor{black}{60.1 $\pm$ 0.4} & \textcolor{black}{26.6 $\pm$ 0.8} & \textcolor{black}{63.9 $\pm$ 0.7} & \textcolor{black}{19.3 $\pm$ 0.4} & \textcolor{black}{45.8 $\pm$ 0.3}           &\textcolor{black}{ 8.8 $\pm$ 0.6}  & \textcolor{black}{83.8 $\pm$ 0.5} & \textcolor{black}{40.9 $\pm$ 0.7} \\
3Player & \textcolor{black}{85.6 $\pm$ 0.6} & \textcolor{black}{29.1 $\pm$ 0.3} & \textcolor{black}{59.3 $\pm$ 0.9} & \textcolor{black}{27.8 $\pm$ 0.5} & \textcolor{black}{64.5 $\pm$ 0.2} & \textcolor{black}{19.6 $\pm$ 0.1} & \textcolor{black}{44.9 $\pm$ 0.6}           & \textcolor{black}{11.9 $\pm$ 0.7} & \textcolor{black}{84.0 $\pm$ 0.4} & \textcolor{black}{40.4 $\pm$ 0.1} \\
              \midrule
              Vanilla Semi-RAT   & 89.8 $\pm$  0.2	&	30.4 $\pm$  0.2	&	63.3 $\pm$  0.4	&	55.4 $\pm$  0.2	&	57.3 $\pm$  0.3	&	43.0 $\pm$  0.1	&	46.1 $\pm$  0.5	&	25.1 $\pm$  0.2 & \textcolor{black}{82.6 $\pm$ 0.6} & \textcolor{black}{40.7 $\pm$ 0.8}\\
              IB (25\% rationales)  & 85.4 $\pm$  0.0	&	28.2 $\pm$  0.0	&	66.4 $\pm$  0.0	&	54.0 $\pm$  0.0	&	63.4 $\pm$  0.0	&	19.2 $\pm$  0.0	&	46.7 $\pm$  0.0	&	10.8 $\pm$  0.0 & \textcolor{black}{88.8 $\pm$ 0.0} & \textcolor{black}{63.9 $\pm$ 0.0} \\
              WSEE        &90.1 $\pm$  0.1	&	32.2 $\pm$  0.1	&	65.0 $\pm$  0.8	&	55.8 $\pm$  0.5	&	59.9 $\pm$  0.4	&	43.6 $\pm$  0.4	&	49.2 $\pm$  0.9	&	14.8 $\pm$  0.8 & \textcolor{black}{84.3 $\pm$ 0.3} & \textcolor{black}{44.9 $\pm$ 0.5} \\ 
              ST-RAT      & 87.0 $\pm$  0.0	&	31.0 $\pm$  0.0	&	-	&	-	&	62.0 $\pm$  0.0	&	51.0 $\pm$  0.0	&	46.0 $\pm$  0.0	&	9.0 $\pm$  0.0 & \textcolor{black}{89.0 $\pm$ 0.0} & \textcolor{black}{\underline{39.0 $\pm$ 0.0}}\\
              \midrule
              Vanilla Sup-RAT  & \underline{93.6 $\pm$  0.3}	&	38.2 $\pm$  0.2	&	63.8 $\pm$  0.2	&	59.4 $\pm$  0.4	&	61.5 $\pm$  0.3	&	51.3 $\pm$  0.2	&	52.3 $\pm$  0.5	&	16.5 $\pm$  0.2 & \textcolor{black}{83.6 $\pm$ 1.4} & \textcolor{black}{68.9 $\pm$ 0.9}\\
              Pipeline       & 86.0 $\pm$  0.0	&	16.2 $\pm$  0.0	&	63.3 $\pm$  0.0	&	41.2 $\pm$  0.0	&	\underline{62.3 $\pm$  0.0}	&	18.4 $\pm$  0.0	&	\underline{70.8 $\pm$  0.0}	&	\underline{54.8 $\pm$  0.0} & \textcolor{black}{87.7 $\pm$ 0.0} & \textcolor{black}{\underline{81.2 $\pm$ 0.0}}\\
              UNIREX & 91.3 $\pm$ 0.4 & 39.8 $\pm$ 0.6 & 65.5 $\pm$ 0.8 & \underline{62.1 $\pm$ 0.2} & 61.9 $\pm$ 0.7 & 51.4 $\pm$ 0.6 & 48.8 $\pm$ 0.3 & 21.3 $\pm$ 0.1 &\textcolor{black}{81.1 $\pm$ 0.8} &\textcolor{black}{70.9 $\pm$ 0.5} \\
              AT-BMC      & 92.9 $\pm$  0.6	& \underline{40.2 $\pm$  0.3}	&	\underline{65.8 $\pm$  0.2}	&	{61.1 $\pm$  0.5}	&	62.1 $\pm$  0.2	&	\underline{52.1 $\pm$  0.2}	&	49.5 $\pm$  0.4	&	18.6 $\pm$  0.3 &  \textcolor{black}{82.3 $\pm$ 0.3} & \textcolor{black}{71.1 $\pm$ 0.6}\\
              \midrule
              $\textmd{SSR}_{unif}$      & 94.3 $\pm$  0.3	&	33.2 $\pm$  0.4	&	62.8 $\pm$  0.3	&	56.2 $\pm$  0.2	&	60.8 $\pm$  0.4	&	47.6 $\pm$  0.5	&	46.8 $\pm$  0.3	&	26.8 $\pm$  0.2 & \textcolor{black}{86.8 $\pm$ 0.9} & \textcolor{black}{46.6 $\pm$ 0.2}\\
              +random DA\;\,             & 90.7 $\pm$  0.3	&	34.5 $\pm$  0.1	&	63.6 $\pm$  0.5	&	56.1 $\pm$  0.3	&	{61.3 $\pm$  0.7}	&	48.3 $\pm$  0.5	&	46.0 $\pm$  0.1	&	33.1 $\pm$  0.2 & \textcolor{black}{87.4 $\pm$ 0.3} & \textcolor{black}{47.6 $\pm$ 0.5}\\
              +semantic DA               & 90.7 $\pm$  0.2	&	35.6 $\pm$  0.2	&	64.7 $\pm$  0.7	&	42.7 $\pm$  0.4	&	58.0 $\pm$  0.3	& {50.2 $\pm$  0.3}	&	48.7 $\pm$  0.2	&	33.5 $\pm$  0.4 & \textcolor{black}{87.9 $\pm$ 0.6} & \textcolor{black}{48.0 $\pm$ 0.8}\\
              +mixed DA$\quad$           & {94.5 $\pm$  0.2}	&	35.1 $\pm$  0.1	&	65.3 $\pm$  0.6	&	40.3 $\pm$  0.5	&	60.4 $\pm$  0.2	&	49.2 $\pm$  0.5	&	47.6 $\pm$  0.1	&	{35.2 $\pm$  0.2}& \textcolor{black}{88.3 $\pm$ 0.3} & \textcolor{black}{48.8 $\pm$ 0.7}\\
              $-$shared $f_{s}$ and $f_{p}$ & 88.3 $\pm$  0.1	&	29.8 $\pm$  0.6	&	60.2 $\pm$  0.3	&	55.7 $\pm$  0.5	&	57.4 $\pm$  0.3	&	43.5 $\pm$  0.2	&	45.6 $\pm$  0.3	&	24.9 $\pm$  0.2 & \textcolor{black}{81.4 $\pm$ 0.3} & \textcolor{black}{39.3 $\pm$ 0.7}\\
              \midrule
              $\textmd{SSR}_{virt}$      & 90.0 $\pm$  0.0	&	34.6 $\pm$  0.2	&	64.2 $\pm$  0.3	&	{57.0 $\pm$  0.2}	&	58.2 $\pm$  0.5	&	43.8 $\pm$  0.3	& {50.4 $\pm$  0.3}	&	31.3 $\pm$  0.4 & \textcolor{black}{87.1 $\pm$ 0.4} &\textcolor{black}{47.0 $\pm$ 0.5}\\
              +random DA\;\,             & 92.8 $\pm$  0.2	&	36.7 $\pm$  0.2	&	65.4 $\pm$  0.2	&	44.3 $\pm$  0.4	&	58.3 $\pm$  0.6	&	47.7 $\pm$  0.3	&	46.5 $\pm$  0.3	&	32.4 $\pm$  0.2  & \textcolor{black}{87.4 $\pm$ 0.3} & \textcolor{black}{47.5 $\pm$ 0.9}\\
              +semantic DA               & 87.6 $\pm$  0.3	&	36.9 $\pm$  0.1	& {66.2 $\pm$  0.5}	&	49.8 $\pm$  0.4	&	61.1 $\pm$  0.3	&	48.8 $\pm$  0.2	&	46.5 $\pm$  0.4	&	31.1 $\pm$  0.2 & \textcolor{black}{{88.9 $\pm$ 0.2}} & \textcolor{black}{{49.0 $\pm$ 0.1}}\\
              +mixed DA$\quad$           & 90.5 $\pm$  0.2	&	{37.4 $\pm$  0.1}	&	64.5 $\pm$  0.6	&	53.1 $\pm$  0.4	&	60.3 $\pm$  0.2	&	49.1 $\pm$  0.1	&	47.1 $\pm$  0.4	&	33.4 $\pm$  0.3  & \textcolor{black}{88.0 $\pm$ 0.4} & \textcolor{black}{48.5 $\pm$ 0.6} \\
              $-$shared $f_{s}$ and $f_{p}$&  87.9 $\pm$  0.5	&	31.9 $\pm$  0.4	&	62.6 $\pm$  0.3	&	55.3 $\pm$  0.2	&	57.6 $\pm$  0.4	&	43.3 $\pm$  0.5	&	48.6 $\pm$  0.2	&	25.8 $\pm$  0.4 & \textcolor{black}{82.3 $\pm$ 0.5} & \textcolor{black}{40.3 $\pm$ 0.6} \\
              $-$shared $W_{a}$ and $W_{p}$    & 88.3 $\pm$  0.3	&	30.4 $\pm$  0.1	&	62.4 $\pm$  0.9	&	54.0 $\pm$  2.1	&	57.5 $\pm$  0.3	&	42.9 $\pm$  0.1	&	45.8 $\pm$  0.4	&	25.0 $\pm$  0.2 & \textcolor{black}{81.8 $\pm$ 0.7} & \textcolor{black}{39.6 $\pm$ 0.7}\\

              \bottomrule
              \hline
          \end{tabular}
          }
      }   
      \label{table_main2-app}
\end{table*}
Besides, since INVRAT \cite{chang2020invariant} is implemented based on the available environments in data and the real environments are unavailable in ERASER, we employ an environment inference method \cite{li2022learning,yue2024www} to partition ERASER into different environments. Specifically, we  first divide each dataset in  ERASER into two environments, and then reproduce INVRAT using pytorch. 

\subsection{\textcolor{black}{Gold Rationale Efficiency}}
\label{c.1.new}
We additionally investigate SSR with data augmentation on the Evidence Inference dataset. From the experimental results, we observe that SSR with data augmentation performs best on the Evidence Inference dataset when the data augmentation percentage is 25\%.

\begin{table}[!htbp]
  \caption{\textcolor{black}{Gold Rationale Efficiency on Evidence Inference.}}
  \renewcommand\arraystretch{.75}
  \setlength{\tabcolsep}{1.5mm}{
      \scalebox{.6}{
          \begin{tabular}{c|cc|cc|cc|cc|cc|cc}
              \hline    
              \toprule
                  \multirow{2}{*}{${SSR}_{unif}$} &\multicolumn{2}{c|}{$k=0$} & \multicolumn{2}{c|}{$k=5$} & \multicolumn{2}{c|}{$k=10$} &\multicolumn{2}{c}{$k=15$}&\multicolumn{2}{c}{$k=20$}&\multicolumn{2}{c}{$k=25$} \\\cmidrule(r){2-13}
                  &Task&Token-F1&Task&Token-F1&Task&Token-F1 &Task&Token-F1&Task&Token-F1&Task&Token-F1 \\ 
              \midrule
              + $k$\% random DA  &
              46.8$\pm$0.3 & 26.8$\pm$0.2 & 47.3$\pm$0.5 & 28.5$\pm$0.3 & 47.7$\pm$0.4 & 29.0$\pm$0.5 & 48.5$\pm$0.6 & 30.3$\pm$0.2 & 49.2$\pm$0.4 & 30.7$\pm$0.7     & 46.0$\pm$0.1 &  33.1$\pm$0.2\\

              + $k$\% semantic DA  &
              46.8$\pm$0.3 & 26.8$\pm$0.2 & 46.6$\pm$0.1 & 28.9$\pm$0.4 & 47.1$\pm$0.3 & 29.3$\pm$0.5 & 47.6$\pm$0.7 & 31.9$\pm$0.4     & 48.0$\pm$0.7 & 32.1$\pm$0.5 & 48.7$\pm$0.2 & 33.5$\pm$0.4 \\

                  \bottomrule
              \hline

          \end{tabular}
          }
}   
      \label{table_daall1}
    \end{table}

From the above experiments, it is found that the optimal data augmentation percentage is different for different datasets. Since our data augmentation method relies on human labeled rationales and most semi-supervised methods assume that the labeled rationales percentage are 25\%, we also use 25\% as our percentage in our practical implementation.

\subsection{\textcolor{black}{Further Results of Data Augmentation}}
\label{appda}
In this section, to validate the effectiveness of the data augmentation methods,
we present more experimental results about baselines implemented with the random and semantic DA in Table \ref{table_daall1} and Table \ref{table_daall2}.
From the results, we can observe that our SSR achieves competitive results, especially in Token F1 with the same amount of data.

\begin{table}[!htbp]
  \caption{Task F1 and Token F1 of selected rationales for the four datasets with random DA. }
  \renewcommand\arraystretch{.75}
  \setlength{\tabcolsep}{1.5mm}{
      \scalebox{.78}{
          \begin{tabular}{c|cc|cc|cc|cc}
              \hline    
              \toprule
                  \multirow{2}{*}{Methods} &\multicolumn{2}{c|}{Movies} & \multicolumn{2}{c|}{MultiRC} & \multicolumn{2}{c|}{BoolQ} &\multicolumn{2}{c}{Evidence Inference} \\\cmidrule(r){2-9}
                  + random DA&Task&Token-F1&Task&Token-F1&Task&Token-F1 &Task&Token-F1 \\ 
              \midrule

              Vanilla Un-RAT  &88.0 $\pm$ 0.4 &28.4 $\pm$ 0.3 &58.4 $\pm$ 0.2 &24.7 $\pm$ 0.3 &62.1 $\pm$ 0.3 &23.5 $\pm$ 0.2 &47.0 $\pm$ 0.4 &10.4 $\pm$ 0.3 \\

              Vanilla Semi-RAT  &90.6 $\pm$ 0.3 &31.6 $\pm$ 0.1 &64.2 $\pm$ 0.4  &56.2 $\pm$ 0.3  &58.9 $\pm$ 0.1 &44.5 $\pm$ 0.3 &45.0 $\pm$ 0.3  &26.0 $\pm$ 0.3 \\

              WSEE  &89.9 $\pm$ 0.4 &33.4 $\pm$ 0.3  &65.3 $\pm$ 0.1 &55.7 $\pm$ 0.3 &61.0 $\pm$ 0.2 &45.5 $\pm$ 0.3 &50.0 $\pm$ 0.3 &18.7 $\pm$ 0.5  \\
              
              \textcolor{black}{Vanilla Sup-RAT} & \textcolor{black}{93.0 $\pm$ 0.4} & \textcolor{black}{39.1 $\pm$ 0.3} & \textcolor{black}{64.4 $\pm$ 0.6} & \textcolor{black}{60.6 $\pm$ 0.2} & \textcolor{black}{62.1 $\pm$ 0.4} & \textcolor{black}{51.9 $\pm$ 0.7} & \textcolor{black}{52.8 $\pm$ 0.2} & \textcolor{black}{18.5 $\pm$ 0.3} \\

              AT-BMC &92.8  $\pm$  0.1 &40.4 $\pm$ 0.3 &66.6  $\pm$ 0.6 &61.8 $\pm$ 0.5 &62.0 $\pm$ 0.1 &52.6 $\pm$ 0.2 &49.5 $\pm$ 0.3 &19.4  $\pm$ 0.6\\
              \midrule
              $\textmd{SSR}_{unif}$ & 90.7 $\pm$  0.3	&	34.5 $\pm$  0.1	&	63.6 $\pm$  0.5	&	56.1 $\pm$  0.3	&	{61.3 $\pm$  0.7}	&	48.3 $\pm$  0.5	&	46.0 $\pm$  0.1	&	33.1 $\pm$  0.2\\
    
              $\textmd{SSR}_{virt}$   & 92.8 $\pm$  0.2	&	36.7 $\pm$  0.2	&	65.4 $\pm$  0.2	&	44.3 $\pm$  0.4	&	58.3 $\pm$  0.6	&	47.7 $\pm$  0.3	&	46.5 $\pm$  0.3	&	32.4 $\pm$  0.2\\
                  \bottomrule
              \hline

          \end{tabular}
          }
}   
      \label{table_daall1}
    \end{table}

\begin{table}[!htbp]
  \caption{Task F1 and Token F1 of selected rationales for the four datasets with semantic DA. }
  \renewcommand\arraystretch{.75}
  \setlength{\tabcolsep}{1.5mm}{
      \scalebox{.78}{
          \begin{tabular}{c|cc|cc|cc|cc}
              \hline    
              \toprule
                  \multirow{2}{*}{Methods} &\multicolumn{2}{c|}{Movies} & \multicolumn{2}{c|}{MultiRC} & \multicolumn{2}{c|}{BoolQ} &\multicolumn{2}{c}{Evidence Inference} \\\cmidrule(r){2-9}
                  + semantic DA &Task&Token-F1&Task&Token-F1&Task&Token-F1 &Task&Token-F1 \\ 
              \midrule

              Vanilla Un-RAT & 88.3 $\pm$ 0.2 & 28.7 $\pm$ 0.5  & 59.0 $\pm$ 0.3 & 25.1 $\pm$ 0.4 & 61.9 $\pm$ 0.6 & 23.7 $\pm$ 0.4 & 47.3 $\pm$ 0.3 & 11.7 $\pm$ 0.4\\

              Vanilla Semi-RAT & 90.1 $\pm$ 0.3 & 31.3 $\pm$ 0.3 &  64.4 $\pm$ 0.1& 56.6 $\pm$ 0.3 & 59.1 $\pm$ 0.4 & 44.4 $\pm$ 0.2 & 46.6 $\pm$ 0.6 & 26.9 $\pm$ 0.5 \\

              WSEE   & 88.9 $\pm$ 0.7 & 33.1 $\pm$ 0.5 & 64.9 $\pm$ 0.3 & 55.9 $\pm$ 0.3 & 60.9 $\pm$ 0.4 & 46.6 $\pm$ 0.6 & 49.7 $\pm$ 0.3 & 20.9 $\pm$ 0.4 \\
              
              \textcolor{black}{Vanilla Sup-RAT} & \textcolor{black}{92.9 $\pm$ 0.3} & \textcolor{black}{39.9 $\pm$ 0.3} & \textcolor{black}{65.1 $\pm$ 0.4} & \textcolor{black}{59.9 $\pm$ 0.7} & \textcolor{black}{62.2 $\pm$ 0.3} & \textcolor{black}{52.3 $\pm$ 0.5} & \textcolor{black}{53.1 $\pm$ 0.1} & \textcolor{black}{19.9 $\pm$ 0.3}\\
              AT-BMC  & 93.2 $\pm$ 0.3 & 40.7 $\pm$ 0.5 & 66.0 $\pm$ 0.6 & 60.9 $\pm$ 0.4 & 62.2 $\pm$ 0.3 &  52.0 $\pm$ 0.1& 50.0 $\pm$ 0.4 & 22.3 $\pm$ 0.6\\
              \midrule
              $\textmd{SSR}_{unif}$  & 90.7 $\pm$  0.2	&	35.6 $\pm$  0.2	&	64.7 $\pm$  0.7	&	42.7 $\pm$  0.4	&	58.0 $\pm$  0.3	& {50.2 $\pm$  0.3}	&	48.7 $\pm$  0.2	&	33.5 $\pm$  0.4\\
    
              $\textmd{SSR}_{virt}$    & 87.6 $\pm$  0.3	&	36.9 $\pm$  0.1	& {66.2 $\pm$  0.5}	&	49.8 $\pm$  0.4	&	61.1 $\pm$  0.3	&	48.8 $\pm$  0.2	&	46.5 $\pm$  0.4	&	31.1 $\pm$  0.2 \\
                  \bottomrule
              \hline

          \end{tabular}
          }
}   
      \label{table_daall2}
    \end{table}

\subsection{\textcolor{black}{Subjective Evaluation}}
\label{appchat}
\textcolor{black}{Following \cite{sha2021learning,yuedare}, we make a human evaluation to evaluate the rationales with three metrics: usefulness, completeness, and fluency.
Specifically, we randomly select 100 samples from the Movie dataset by comparing ${SSR}_{virt}$ with Inter\_RAT and WSEE.  From the observation on Table \ref{table_human2}, we can find ${SSR}_{virt}$ outperforms Inter\_RAT and WSEE in all metrics, illustrating the effectiveness of ${SSR}_{virt}$.}

\begin{table}[!htbp]
  \caption{\textcolor{black}{Human evaluation on Movies dataset.}}
  \centering
  \renewcommand\arraystretch{1.2}
  \setlength{\tabcolsep}{2.1mm}{
  \scalebox{0.9}{
        \begin{tabular}{c|ccc}
           \hline    
           \toprule
               
              Methods&Usefulness&Completeness&Fluency\\
           \midrule
           Inter\_RAT       & 3.69 & 3.53 & 3.88 \\
           WSEE    & 3.82 & 3.78 & 4.05 \\
              \midrule
              ${SSR}_{virt}$        & {\bf3.90} & {\bf3.88} & {\bf4.20} \\
           \bottomrule
           \hline
        \end{tabular}
     }
  }   
  \label{table_human2}
\end{table}

Considering the remarkable success of Large Language Models, we employ ChatGPT \citep{ChatGPT} as an alternative to humans for rationale evaluation. 
\textcolor{black}{From the observation on Table \ref{table_human}, we find that our model achieves a similar partial order relationship between human evaluation and ChatGPT evaluation. This further illustrates the effectiveness of our method.}

\begin{table}[!htbp]
   \caption{ChatGPT evaluation on Movies dataset.}
   \centering
   \renewcommand\arraystretch{1.2}
   \setlength{\tabcolsep}{2.1mm}{
   \scalebox{0.9}{
         \begin{tabular}{c|ccc}
            \hline    
            \toprule
                
               Methods&Usefulness&Completeness&Fluency\\
            \midrule
            Inter\_RAT       & 3.80 & 3.76 & 4.02 \\
            WSEE    & 3.86 & 3.88 & 3.96 \\
               \midrule
               ${SSR}_{virt}$        & {\bf4.06} & {\bf3.94} & {\bf4.13} \\
            \bottomrule
            \hline
         \end{tabular}
      }
   }   
   \label{table_human}
\end{table}

This is our prompt to ChatGPT:

Now, you are an annotator. First, I'll give you some original movie review text with labels to the text (i.e., positive or negative). After that, I will give you some rationales (i.e., one or more consecutive sentences and tokens extracted from the text) generated by a neural network, as well as its predicted labels that judge whether the movie review is positive or negative. Finally, you are asked to evaluate these extracted rationales from three metrics including Usefulness, Completeness and Fluency. Among them, each metric is from 1 (lowest) to 5 (e.g. 3.4 and 4.0).

Detailed standards for annotators:

\textbf{Usefulness:}

Q: Do you think the selected rationales can be useful for explaining the predicted labels?

$\bullet$ 5: Exactly. Selected rationales are useful for me to get the correct label.

$\bullet$ 4: Highly useful. Although several tokens have no relevance to correct label, most selected tokens are useful to explain the labels.

$\bullet$ 3: Half of them are useful. About half of the tokens are useful for getting labels.

$\bullet$ 2: Almost useless. Almost all of tokens are useless.

$\bullet$ 1: No Use. The selected rationales are useless for identifying labels.

\textbf{Completeness:}

Q: Do you think the selected rationales are enough for explaining the predicted labels?

$\bullet$ 5: Exactly. Selected rationales are enough for me to get the correct label.

$\bullet$ 4: Highly complete. Several tokens related to the label are missing.

$\bullet$ 3: Half complete. There are still some important tokens that have not been selected, and they are in nearly the same number as the selected tokens.

$\bullet$ 2: Somewhat complete. The selected tokens are not enough.

$\bullet$ 1: Nonsense. None of the important tokens is selected.

\textbf{Fluency:}

Q: Do you think the selected rationales are fluent?

$\bullet$ 5: Very fluent.

$\bullet$ 4: Highly fluent.

$\bullet$ 3: Partial fluent.

$\bullet$ 2: Very unfluent.

$\bullet$ 1: Nonsense.

\subsection{\textcolor{black}{Other metrics}}
\label{c.4.new}
\textcolor{black}{We use the  comprehensiveness and sufficiency metrics from ERASER \citep{DeYoungJRLXSW20}, and report the corresponding results in the table. As shown in Table \ref{othermetric}, ${SSR}_{unif}$ and  ${SSR}_{virt}$ still performs  better than baselines.}

\begin{table*}[!htbp]
  \caption{\textcolor{black}{Comprehensiveness and sufficiency of selected rationales.} }
  \vspace{-0.3cm}
  \centering
  \renewcommand\arraystretch{.75}
  \setlength{\tabcolsep}{1.5mm}{
      \scalebox{.78}{
          \begin{tabular}{c|cc|cc}
              \hline    
              \toprule
                  \multirow{2}{*}{Methods} &\multicolumn{2}{c|}{IMDB} & \multicolumn{2}{c}{SST-2} \\\cmidrule(r){2-5}
                  &suff ($\downarrow$)&com ($\uparrow$)&suff ($\downarrow$)&com ($\uparrow$) \\ 
              \midrule
              Vanilla Un-RAT   & 0.22                & 0.33             & 0.38                & 0.18 \\
              Vanilla Semi-RAT & 0.15                & 0.40             & 0.24                & 0.23   \\
              WSEE             & 0.13                & 0.42             & 0.20                & 0.26  \\
              \midrule
              ${SSR}_{unif}$   & 0.10                & 0.41             & 0.14                & 0.32  \\
              ${SSR}_{virt}$   & \textbf{0.07}                & \textbf{0.45}             & \textbf{0.11}                & \textbf{0.35}  \\
                  \bottomrule
              \hline

          \end{tabular}
          }
}   
      \label{othermetric}
       \vspace{-0.4cm}
    \end{table*}

\subsection{Visualized Selective Rationales}
\label{appvis}
\begin{figure}[!htp]
  \centering
  \includegraphics[width=14cm]{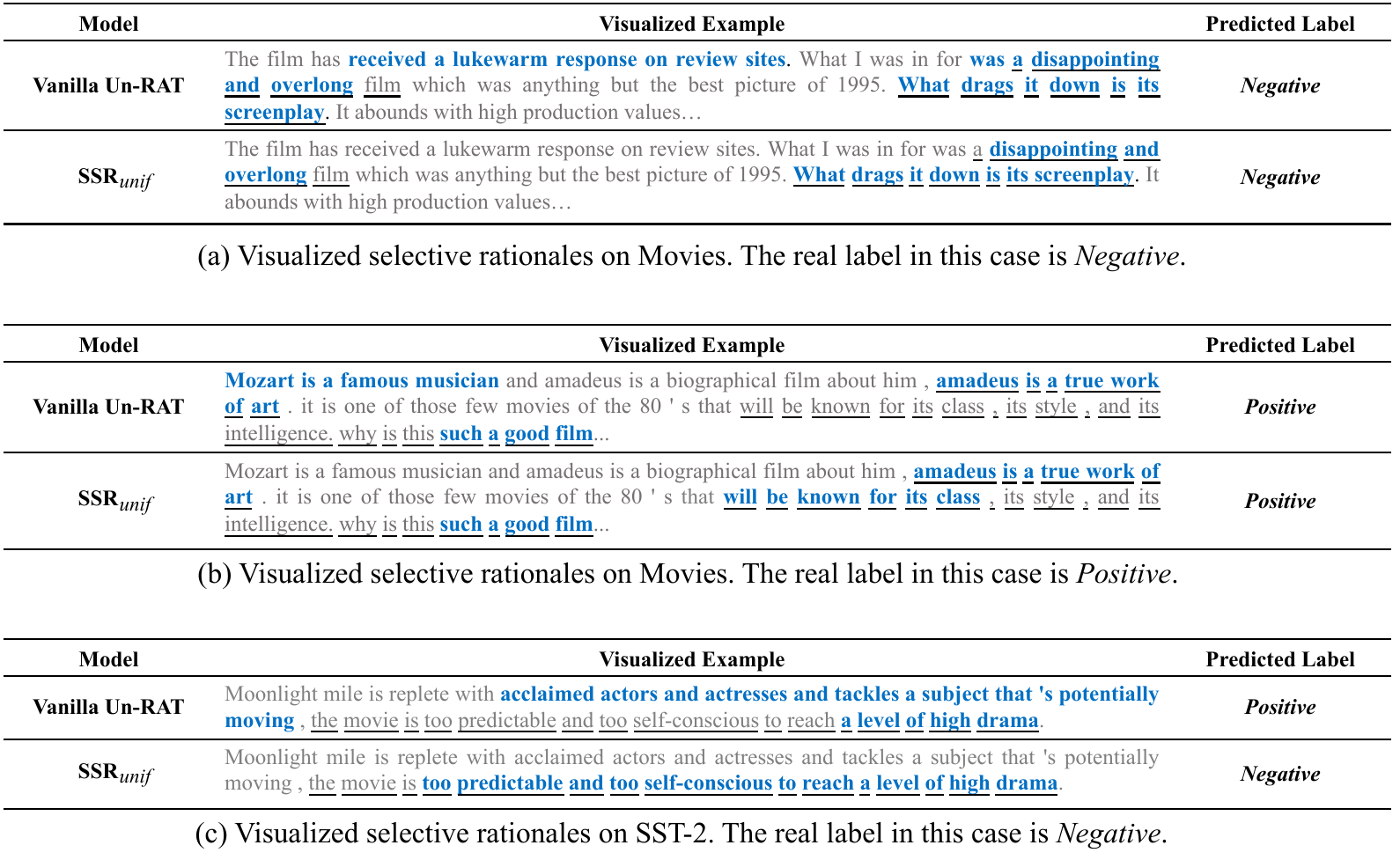}
  \caption{
    A visualized performance of extracted rationales with different methods. 
  } 
  \label{case}
\end{figure}

In this section, we provide more visualization cases in Figure \ref{case} to show the performance of select rationales. From the observation, we can find that SSR can select faithful rationales.

Specifically,  we show three examples of rationales selected by Vanilla Un-RAT and our $\textmd{SSR}_{unif}$ on both Movies and SST-2, where the label is \textit{Negative} or \textit{Positive}.
Among them, the examples in Figure \hyperref[case]{6(a)} and \hyperref[case]{6(b)} come from Movies and the example in  Figure \hyperref[case]{6(c)} comes from SST-2. The underlined tokens represent the ground truth rationales, and the \hlgreen1{blue} is the predicted rationales.

In Figure \hyperref[case]{6(a)}, where the label is \textit{Negative}, we find although both Vanilla Un-RAT and $\textmd{SSR}_{unif}$ predict the label as \textit{Negative} correctly, Vanilla Un-RAT still extracts shortcuts as rationales. Specifically,  ``received a lukewarm response'' is the shortcuts, where human being judges a movie is not influenced by other reviews. $\textmd{SSR}_{unif}$ avoids these shortcuts, but Vanilla Un-RAT extracts these as rationales.

In Figure \hyperref[case]{6(b)}, where the label is \textit{Positive}, we observe find although both Vanilla Un-RAT and $\textmd{SSR}_{unif}$ predict the label as \textit{Positive} correctly, Vanilla Un-RAT still extracts shortcuts as rationales. Specifically, ``Mozart is a famous musician'' is the shortcuts. Although Mozart was a great musician, it has no relevance to how good his biographical film is. Our $\textmd{SSR}_{unif}$  avoids these shortcuts, but Vanilla Un-RAT extracts these as rationales.

In Figure \hyperref[case]{6(c)}, where the label is \textit{Negative}, we can find $\textmd{SSR}_{unif}$ predicts the label as \textit{Negative} correctly but Vanilla Un-RAT fails. Specifically, since Vanilla Un-RAT relies on shortcuts in the data for prediction, when Vanilla Un-RAT is generalized to the OOD dataset (i.e., SST-2), the task performance decreases due to the changed data distribution. And some wrong rationales are extracted. Our $\textmd{SSR}_{unif}$, on the other hand, extracts the rationales accurately and predicts the task results correctly, indicating the effectiveness of exploring shortcuts to predict task results.

\section{Discussions of  SSR and LLMs}
Since SSR can mitigate the problem of utilizing shortcuts to compose rationales, our work can be applied to certain decision-making domains such as the judicial domain and the medical domain. 
In addition, although Large Language Models (LLMs) has achieved remarkable results recently, most LLMs (e.g. Chatgpt \citep{ouyang2022training,ChatGPT} and Claude \citep{Claude}) need to be called as APIs. It cannot be used in privacy and security critical systems such as the judicial system. Our approach is much easier to deploy locally, ensuring privacy and improving explainability.
Meanwhile,  SSR is a model-agnostic approach, and we can replace BERT in SSR with opened LLMs (e.g., LLaMA \citep{touvron2023llama}) to achieve high performance. 

SSR offers a solution for offering explanations in text classification tasks. 
For LLMs, in the autoregressive generation,  we can view the prediction of each token as a text classification task, with the number of categories matching the vocabulary size. By generating an explanation for each token, we can gradually explain the output of a LLM.
Nevertheless, effectively applying SSR to explain LLMs remains a formidable challenge. We are committed to further researching and exploring this area as a primary focus of our future work.


\end{document}